%% file: neurips_2026.tex
\documentclass{article}

\usepackage[main, final]{neurips_2026}
\usepackage[utf8]{inputenc} % allow utf-8 input
\usepackage[T1]{fontenc}    % use 8-bit T1 fonts
\usepackage{hyperref}       % hyperlinks
\usepackage{url}            % simple URL typesetting
\usepackage{booktabs}       % professional-quality tables
\usepackage{amsfonts}       % blackboard math symbols
\usepackage{nicefrac}       % compact symbols for 1/2, etc.
\usepackage{microtype}      % microtypography
\usepackage{xcolor}         % colors
\usepackage{hyperref}
\usepackage{tabularx}
\usepackage{graphicx}
\usepackage{amsmath}
\usepackage{amssymb}
\usepackage{float}
\usepackage{wrapfig}
\usepackage{fvextra}
\usepackage{enumitem}
\usepackage{soul}
\soulregister\cite7   % 让 \cite 能在 \hl 里正常工作
\soulregister\ref7
\soulregister\pageref7

\usepackage{tcolorbox}
\tcbuselibrary{most}

\usepackage{tcolorbox}

\usepackage{tcolorbox}
\usepackage[table]{xcolor}   % 自带 colortbl 功能

\usepackage{xcolor}
\newcommand{\qualitativeSample}[5]{
\begin{tcolorbox}[colback=white, colframe=gray!50, arc=2mm, boxrule=0.5pt, left=2mm, right=2mm, top=2mm, bottom=2mm]
    \centering
    \includegraphics[height=4.5cm, keepaspectratio]{#1}
    
    \tcbline
    \begin{flushleft}
        \small
        \textbf{\textsf{Qwen3-vl-4b-instruct (Untrained):}}\par
        \textbf{CoT:} #2\par
        \vspace{1mm}
        \textbf{Prediction:} \textcolor{red}{\textbf{#3}} \texttimes
    \end{flushleft}
    
    \tcbline
    \begin{flushleft}
        \small
        \textbf{\textsf{Ours :}}\par
        \textbf{CoT:} #4\par
        \vspace{1mm}
        \textbf{Prediction:} \textcolor{blue}{\textbf{#5}} \checkmark
    \end{flushleft}
\end{tcolorbox}
}

\newcommand{\badcaseSample}[4]{
\begin{tcolorbox}[colback=white, colframe=gray!50, arc=2mm, boxrule=0.5pt, left=2mm, right=2mm, top=2mm, bottom=2mm]
    \centering
        \begin{flushleft}
        \small
        \textbf{\textsf{Bad case}}\par

    \end{flushleft}
    \includegraphics[height=4.5cm, keepaspectratio]{#1}
    
    \tcbline
    \begin{flushleft}
        \small
        \textbf{\textsf{Ours:}}\par
        \textbf{CoT:} #2\par
        \vspace{1mm}
        \textbf{Prediction:} \textcolor{red}{\textbf{#3}} \texttimes\par
        \textbf{GT Prediction:} \textbf{#4}

    \end{flushleft}
    
\end{tcolorbox}
}

\title{Reasoning-Aware Relational Representation Learning for Open-Vocabulary Scene Graph Generation}

\author{
Jiawei~Xiao$^1$\thanks{Equal contribution.}  , Boya Wang$^1$$^*$,Longtian Qiu$^1$,Shan Ning$^{1,3}$, Xuming He$^{1,2}$\thanks{ Corresponding author.} \\
$^1$ShanghaiTech University, Shanghai, China\\
$^2$Shanghai Engineering Research Center of Intelligent Vision and Imaging\\
$^3$Lingang Laboratory, Shanghai, China\\
\texttt{\{xiaojw2025, wangby2025, qiuilt, ningshan2022, hexm\}@shanghaitech.edu.cn}
}

\begin{document}

\maketitle

% \begin{abstract}
%   The abstract paragraph should be indented \nicefrac{1}{2}~inch (3~picas) on both the left- and right-hand margins. Use 10~point type, with a vertical spacing (leading) of 11~points. The word \textbf{Abstract} must be centered, bold, and in point size 12. Two line spaces precede the abstract. The abstract must be limited to one paragraph.
% \end{abstract}

% \section{Introduction}

% \section{Related Work}
% \paragraph{MLLMs}
% \paragraph{SGG}

% %通过介绍mllm，引入我们使用mllm生成cot作为我们的表征
% \paragraph{Open-Vocabulary SGG (OVSGG)}
% %介绍先前的ovsgg工作

% \section{method}

% \section{Experiments}
% \subsection{Experiment Setup}
% %介绍测试的数据集，介绍三种setting，评价指标，着重强调mr的作用
% \subsection{Comparisons with OVR-SGG Methods}
% \paragraph{Predcls}

% \paragraph{Sgcls}
% \paragraph{SGDet}

% \subsection{Diagnostic Experiment}

% \paragraph{Ablation Study}

% \section{Conclusion}

\input{sections/abstract}
\input{sections/intro}

\input{sections/related_works}

\input{sections/methods}

\input{sections/exps}
\input{sections/conclusion}
\input{sections/acknowledge}

\bibliographystyle{plain}
\bibliography{sections/reference.bib}

%%%%%%%%%%%%%%%%%%%%%%%%%%%%%%%%%%%%%%%%%%%%%%%%%%%%%%%%%%%%

\appendix

\input{sections/appendix}

%%%%%%%%%%%%%%%%%%%%%%%%%%%%%%%%%%%%%%%%%%%%%%%%%%%%%%%%%%%%

\newpage
\input{sections/checklist}

\end{document}

%% file: sections/abstract.tex
\begin{abstract}

Open-Vocabulary Scene Graph Generation (OV-SGG) requires models to recognize visual relationships beyond the training vocabulary, yet existing methods often rely on dataset-specific object--predicate co-occurrence patterns. Under long-tailed distributions and noisy object localization, such reliance leads to biased, entangled relation representations, limiting both rare-relation recognition and generalization to unseen predicates.
To address these challenges, we propose RaRe, a reasoning-aware relational representation learning framework that transforms MLLM-generated chain-of-thought (CoT) rationales into structured relation embeddings. Rather than using CoT only as an intermediate reasoning trace for final prediction, RaRe directly optimizes the rationale-derived representation space to improve inter-relation discriminability. Specifically, RaRe first aligns rationale embeddings with textualized relation descriptions through self-supervised contrastive learning, and then refines rationale generation with a specificity-oriented reinforcement learning objective that encourages semantically object-pair-specific relational evidence.
Experiments on VG150 and PSG demonstrate improved category-balanced recognition over existing baselines, while the additional gains from learning vary across datasets and proposal settings. Code available at: \url{https://artanic30.github.io/project_pages/RaRe}.

\end{abstract}

%% file: sections/intro.tex
\section{Introduction}

Scene Graph Generation (SGG) represents an image as a structured graph of objects and their pairwise relationships, providing a compact abstraction that supports structured reasoning over visual content~\citep{wang2018scene}. This representation has been widely adopted in vision-language tasks such as visual question answering~\citep{shi2019explainable,hudson2019gqa}, image captioning~\citep{gu2019unpaired,yang2019auto}, and image retrieval~\citep{Johnson_2015_CVPR}. While early SGG methods typically operated under a closed-set assumption with a fixed predicate vocabulary, recent work has shifted toward Open-Vocabulary Scene Graph Generation (OV-SGG), which enables models to generalize beyond the training vocabulary by capturing transferable relational semantics over novel object compositions and unseen predicates.

\begin{figure}[!t]
  \centering
  \includegraphics[width=\linewidth]{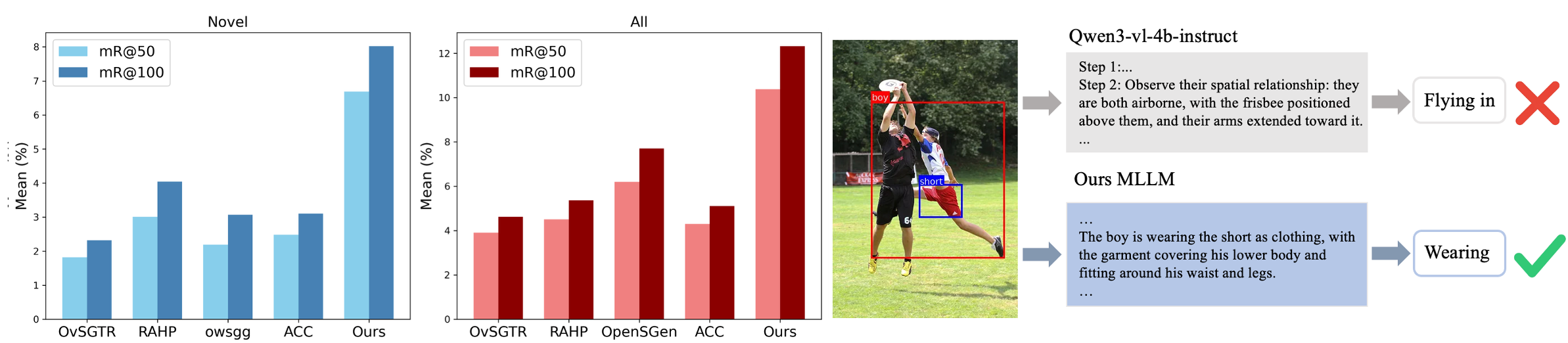}
 \vspace{-1mm}
   \caption{Left: our method achieves higher mR@50/100 on both novel and all predicate categories compared with existing OV-SGG methods on SGDet. Right: our method generates more object-pair-specific CoT relational evidence, leading to a better relation prediction than the baseline MLLM.}
    \label{fig:teaser}
    \vspace{-6mm}
\end{figure}

%Existing OV-SGG approaches predominantly fall into two paradigms. \textit{Embedding-matching methods}~\citep{he2022towards,yu2023visuallyEPIC,chen2024expanding,li2025interaction} leverage vision-language alignment to transfer open-vocabulary recognition capabilities to relation prediction. Meanwhile, \textit{generative methods}~\citep{li2024pixelsgraphsopenvocabularyscene,zhou2024openpsgopensetpanopticscene,xu2024llavaspacesggvisualinstructtuning} formulate the task as a sequence generation problem, using generative models to directly output relational descriptions in natural language. Despite their progress, both paradigms still face three main challenges. First, their generalization to novel relations can be further improved. Performance on unseen predicates drops precipitously compared to seen predicates, indicating that these models tend to overfit to dataset-specific co-occurrence biases rather than acquiring genuinely transferable relational semantics. Second, maintaining robustness under the realistic Scene Graph Detection setting remains difficult. Relation prediction performance tends to degrade when upstream object detections are noisy, indicating a sensitivity to input perturbations. Third, class imbalance under long-tailed distributions continues to be an open issue. While existing methods achieve competitive overall recall (R@K), their mean recall (mR@K) is relatively lower, reflecting the difficulty in recognizing rare relations within imbalanced datasets.

Existing OV-SGG approaches can be broadly categorized into two paradigms. \textit{Embedding-matching methods}~\citep{he2022towards,yu2023visuallyEPIC,chen2024expanding,li2025interaction} leverage vision-language alignment to transfer open-vocabulary recognition capabilities to relation prediction. In contrast, \textit{generative methods}~\citep{li2024pixelsgraphsopenvocabularyscene,zhou2024openpsgopensetpanopticscene,xu2024llavaspacesggvisualinstructtuning} cast the task as sequence generation, directly producing relational descriptions in natural language. Despite their progress, both paradigms exhibit several persistent limitations. First, generalization to unseen predicates remains limited, with performance often dropping significantly compared to seen relations, suggesting that current models tend to rely on dataset-specific object–predicate co-occurrence patterns rather than learning transferable relational semantics. Second, robustness under the realistic Scene Graph Detection (SGDet) setting is still inadequate. Relation prediction is highly sensitive to object localization errors, as inaccurate object candidate pairs can degrade relational reasoning. Third, long-tailed data distributions pose a more fundamental challenge in OV-SGG, as training is dominated by frequent object–predicate co-occurrence patterns, leading to biased and entangled representations that collapse toward frequent predicates, limiting both tail discrimination among seen classes and generalization to unseen ones. Consequently, models exhibit degraded performance on rare relations and limited transferability beyond the training distribution, as reflected in the persistent gap between overall recall (R@K) and mean recall (mR@K).

To address these challenges, we revisit OV-SGG from the perspective of relational representation learning. Rather than focusing on how to better utilize MLLMs for prediction, we treat them as a source of structured relational signals. Specifically, we propose a \textit{reasoning-aware} relational representation learning framework that extracts instance-level rationales from MLLMs—such as spatial configurations, object attributes, and action states—and encodes them into structured embeddings for downstream relation prediction. By grounding relation representations in these compositional cues, our approach mitigates reliance on dataset-specific object–predicate co-occurrence patterns and promotes more transferable relational semantics. 
A key challenge, however, lies in the nature of these rationale signals: since MLLMs are trained with generative objectives that prioritize linguistic plausibility, the resulting representations are not optimized for inter-class discriminability. This often leads to entangled embeddings for semantically similar relations and exacerbates representation collapse under long-tailed distributions. To address this, our framework combines contrastive alignment of rationale--prediction pairs with specificity-driven policy optimization to learn relation-discriminative representations for open-vocabulary retrieval.

Concretely, we adopt a two-stage learning strategy. In the first stage, we employ self-supervised alignment to explicitly match MLLM-generated chain-of-thought (CoT) rationales with textualized relation descriptions, constructing a unified semantic space that captures fine-grained relational distinctions. In the second stage, we introduce reinforcement learning to refine the rationale generation process, encouraging the model to produce CoT outputs with clearer semantic separation across relation categories. Through this two-stage optimization process, the framework progressively learns more stable, discriminative, and transferable relational representations.

%Experiments on the Visual Genome (VG150) demonstrate that our approach significantly enhances OV-SGG performance in multiple critical dimensions. Specifically, results underscore the multifaceted advantages of our approach: it first secures superior generalization with a 22.49 Novel mR@50, while simultaneously maintaining a balanced recognition on base categories (25.53 mR@50) in predcls. More importantly, we maintained a solid 9.63 mR@50 even in the noise-prone SGDet setting. These findings validate that CoT-guided relational representations effectively address the core limitations of existing paradigms, offering a promising new perspective for open-vocabulary scene understanding.

Experiments on Visual Genome (VG150) demonstrate that our approach improves OV-SGG performance across several key dimensions. In particular, the model achieves strong generalization to novel predicates (22.49 Novel mR@50) while maintaining balanced recognition on base categories (25.53 mR@50) under the PredCls setting. Under SGDet, RaRe achieves 10.38 Total mR@50 on VG150 and 17.52 on PSG using detector-generated proposals. Ablations further distinguish the contributions of the underlying retrieval pipeline from the setting-dependent gains of representation learning and calibration.

Our main contributions are summarized as follows:
\vspace{-10pt}

\begin{itemize}[leftmargin=5mm]
    \item We propose a reasoning-aware relational representation learning framework for OV-SGG that addresses challenges in novel relation generalization, robustness of predicate recognition under noisy localization, and long-tailed distributions.
    \item We introduce structured MLLM-generated rationales as instance-level semantic priors for relational representation, together with a two-stage training strategy that integrates self-supervised alignment and reinforcement learning to extract and refine discriminative relational cues.
    \item Experiments validate that our method achieves consistent and notable improvements over existing approaches across open-vocabulary generalization tasks, realistic noisy detection (SGDet) settings, and long-tailed evaluation metrics.
\end{itemize}

%% file: sections/related_works.tex
\section{Related Work}

\subsection{Scene Graph Generation}
Scene Graph Generation (SGG) aims to localize objects and predict pairwise visual relationships as subject–predicate–object triplets in images~\cite{krishna2017visual}. Most existing methods formulate predicate prediction as a classification problem over a predefined set of categories~\cite{xu2017scene,Zellers_2018_CVPR}. However, real-world data exhibits a long-tailed distribution of predicates, where frequent relations dominate training and rare ones are underrepresented, leading to biased predictions~\cite{Tang_2020_CVPR}. Moreover, these methods are limited to a fixed set of predefined categories and do not address generalization to unseen relations.

Open-Vocabulary SGG (OV-SGG)~\cite{he2022towards} extends conventional SGG by requiring models to recognize predicates unseen during training. Embedding-based methods leverage vision-language models such as CLIP~\cite{radford2021learningtransferablevisualmodels} to align visual relation features with textual semantic embeddings~\cite{VS3,chen2024expanding}, with recent works further enhancing textual representations or incorporating interaction-aware prompting strategies~\cite{yu2023visuallyEPIC,li2025interaction,liu2025relationawarehierarchicalpromptopenvocabulary} to improve novel predicate recognition. Another line of work adopts generative formulations: early approaches rely on constrained decoding guided by vision-language alignment~\cite{he2022towards,li2024pixelsgraphsopenvocabularyscene}, while more recent methods employ multimodal large language models (MLLMs) to directly produce scene graphs as structured natural language~\cite{zhou2024openpsgopensetpanopticscene,xu2024llavaspacesggvisualinstructtuning,openSGen}. Beyond the open-vocabulary setting, LLM4SGG~\cite{kim2024llm4sgglargelanguagemodels} uses LLMs to augment relational supervision, and Relation-R1~\cite{li2025relationr1progressivelycognitivechainofthought} introduces CoT reasoning with reinforcement learning to boost their performance under open-ended settings. Despite their progress, existing methods generally optimize for generation accuracy or embedding alignment, without explicitly enforcing semantic separability across predicates—particularly under long-tailed distributions and noisy detection conditions. Our work bridges the embedding-based and generative paradigms by constructing a discriminative relational embedding space and using it to drive the MLLM's generative policy toward instance-specific outputs.

\subsection{Chain-of-Thought Reasoning in Multimodal LLMs}
Chain-of-Thought (CoT) prompting~\cite{wei2023chainofthoughtpromptingelicitsreasoning} enables step-by-step reasoning in LLMs and has been extended to multimodal settings~\cite{zhang2024multimodalchainofthoughtreasoninglanguage}. In visual relation understanding, Relation-R1~\cite{li2025relationr1progressivelycognitivechainofthought} applies CoT reasoning for SGG, demonstrating that explicit intermediate reasoning can well enhance prediction quality. While this approach utilizes reinforcement learning to optimize CoT paths, its reward is defined at the prediction level, primarily based on the correctness of final outputs. Consequently, CoT is leveraged to guide reasoning and prediction, but is not explicitly modeled within a representation space where its semantic structure can be directly optimized. In contrast, we treat CoT rationales as relational representations and learn an embedding space to optimize their semantic specificity.

\subsection{Reinforcement Learning and Contrastive Learning for LLMs}
% Longtian version - Reinforcement learning has emerged as a powerful tool for aligning LLM generation with task-specific objectives. Under the framework of Reinforcement Learning from Human Feedback (RLHF)~\cite{ouyang2022traininglanguagemodelsfollow}, variants such as DPO~\cite{rafailov2023direct} align outputs with human preferences, while Group Relative Policy Optimization (GRPO)~\cite{shao2024deepseekmathpushinglimitsmathematical,Guo_2025} enables efficient policy optimization without a separate critic model.

Reinforcement learning (RL) has emerged as a powerful tool for aligning LLM generation with task-specific objectives.
Reinforcement Learning from Human Feedback (RLHF)~\cite{ouyang2022traininglanguagemodelsfollow} instantiates this paradigm by deriving reward signals from human preferences.
Group Relative Policy Optimization (GRPO)~\cite{shao2024deepseekmathpushinglimitsmathematical,Guo_2025} is a policy optimization method that has been widely adopted for LLM post-training, where advantages are estimated via group-relative reward normalization, removing the need for a learned value network.
These methods have achieved success in mathematical reasoning and instruction following, but their reward designs typically rely on ground-truth verification or human annotation, limiting applicability to tasks where labels are scarce or ambiguous.

In parallel, contrastive and self-supervised representation learning~\cite{chen2020simpleframeworkcontrastivelearning,he2020momentumcontrastunsupervisedvisual,oord2019representationlearningcontrastivepredictive,wang2020understanding,haochen2021provable} has proven effective at encouraging inter-class separability in embedding spaces, underpinning many vision-language systems~\cite{radford2021learningtransferablevisualmodels} and text embedding methods~\cite{gao2021simcse}. % [需要确认cite key] SimCSE 已确认
While effective for representation quality, these methods operate independently from the generative process of language models.

Our work unifies these two directions: we first learn a discriminative embedding space via self-supervised contrastive learning over CoT reasoning-prediction pairs, and then derive a label-free specificity reward from this space to optimize the MLLM's generation policy via GRPO—closing the loop between representation discriminability and generation quality.

%% file: sections/methods.tex
\section{Method}

\subsection{Preliminaries and Task Definition}

\paragraph{Open-Vocabulary Scene Graph Generation.}
Given an image $I$, Scene Graph Generation (SGG) aims to produce a scene graph $\mathcal{G} = \{\mathcal{V}, \mathcal{R}\}$, where $\mathcal{V} = \{v_i\}_{i=1}^{N}$ is a set of entities with $v_i = (c_i^v, b_i)$ comprising a category label $c_i^v \in \mathcal{C}^v$ and a bounding box $b_i \in \mathbb{R}^4$, and $\mathcal{R} = \{r_{ij}\}_{i \neq j}$ is a set of relationships, each represented as a triplet $r_{ij} = \langle v_i, p_{ij}, v_j \rangle$ with predicate $p_{ij} \in \mathcal{C}^p$. In the \textit{open-vocabulary} setting, $\mathcal{C}^p$ is split into base classes $\mathcal{C}^p_\text{base}$ and novel classes $\mathcal{C}^p_\text{novel}$ ($\mathcal{C}^p_\text{base} \cap \mathcal{C}^p_\text{novel} = \emptyset$). During training, only annotations from $\mathcal{C}^p_\text{base}$ are available, while at test time the model is required to generalize to both base and novel predicates.

\paragraph{MLLM-based Relation Prediction.}
We formulate predicate prediction as conditional text generation. Given an image $I$ and a prompt $\mathbf{x}$ specifying the entity pair $(v_i, v_j)$, the MLLM autoregressively generates a response $\mathbf{y} = (y_1, \ldots, y_T)$:
\begin{equation}
    p_\theta(\mathbf{y} \mid I, \mathbf{x}) = \prod_{t=1}^{T} p_\theta(y_t \mid I, \mathbf{x}, \mathbf{y}_{<t}),
\end{equation}
where $\theta$ denotes the MLLM parameters. The output consists of a chain-of-thought reasoning followed by a predicate prediction, denoted as $\mathbf{y} = [\mathbf{y}^\text{cot}; \mathbf{y}^\text{pred}]$.

%To bridge the MLLM's free-form textual outputs and the structured relation scores required by SGG, we employ an embedding model $f_\phi$ that maps a text sequence $\mathbf{s}$ to a normalized vector $\mathbf{e} = f_\phi(\mathbf{s}) \in \mathbb{R}^d$ ($\|\mathbf{e}\| = 1$). Given relation category prototypes $\{\mathbf{c}_k\}_{k=1}^{|\mathcal{C}^p|}$ in the same space, the relation score for category $k$ is:
%\begin{equation}
%    s_k = \frac{\mathbf{e}^\top \mathbf{c}_k}{\|\mathbf{e}\| \|\mathbf{c}_k\|}.
%\end{equation}

\textbf{Relation Scoring via Embedding Matching.}
To convert the MLLM's free-form response into structured predicate scores, we employ an embedding model $f_\phi$ that maps a text sequence to a normalized vector. 
Given the generated response $y$, we obtain its relation representation as $\mathbf{e}=f_\phi(y)\in\mathbb{R}^d$ with $\|\mathbf{e}\|=1$. 
For each predicate category $k$, we construct a textual prototype $\mathbf{c}_k$ in the same embedding space. 
The relation score for category $k$ is computed by cosine similarity:
\begin{equation}
    s_k = \frac{\mathbf{e}^\top \mathbf{c}_k}{\|\mathbf{e}\|\|\mathbf{c}_k\|}.
\end{equation}
This embedding-based scoring bridges CoT-style generation and structured relation prediction, and is later instantiated with multi-template predicate prototypes in Sec.~\ref{sec:inference}.

\begin{figure*}[t!]
  \centering
  \includegraphics[width=1.0\textwidth]{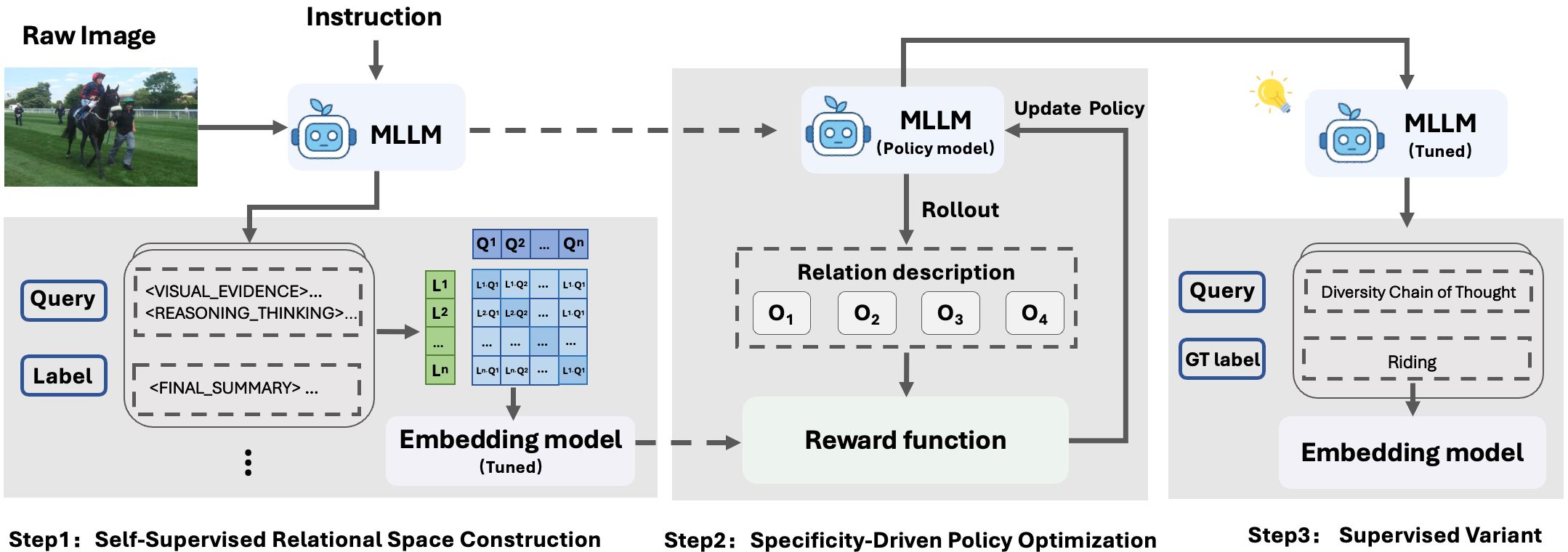}

   \caption{\textbf{Training pipeline of our method.} Step 1 we introduce Self-supervised training for the reward model. Step 2 we adopt GRPO for optimizing the MLLM. Step 3 we adopt supervised tuning for the final retrieval model.}
    \label{fig:pipeline}
     \vspace{-4mm}
\end{figure*}

\subsection{Overview}
% To address the challenges of generalization to novel relations, robustness under noisy detections, and long-tailed bias in open-vocabulary scene graph generation, we formulate the problem as relational representation learning 
% We propose a two-stage framework for open-vocabulary scene graph generation that learns transferable and stable relational semantics beyond training distributions. As shown in Fig.~\ref{fig:pipeline}, our framework consists of two core stages, followed by an optional supervised retrieval tuning step. In Stage 1, we construct a transferable relational embedding space via self-supervised contrastive learning by exploiting the intrinsic pairing between chain-of-thought (CoT) reasoning and relation predictions, enabling instance-level relational representations without explicit annotations and reducing dataset bias(Sec.~\ref{sec:stage1}). In Stage 2, we optimize the MLLM's generation policy through reinforcement learning driven by a specificity reward derived from this embedding space, encouraging instance-grounded and semantically distinctive reasoning to improve robustness and generalization(Sec.~\ref{sec:stage2}). Overall, the embedding space learned in Stage 1 serves as a bridge between free-form reasoning and structured relation prediction, and provides guidance for optimizing the generation process in Stage 2.
We propose a two-stage framework for open-vocabulary scene graph generation that learns transferable and stable relational semantics beyond training distributions. As shown in Fig.~\ref{fig:pipeline}, our framework consists of two core stages, followed by an optional supervised retrieval tuning step. In Stage 1, we learn a relational embedding space via contrastive learning by leveraging the intrinsic pairing between chain-of-thought (CoT) reasoning and relation predictions (Sec.~\ref{sec:stage1}). Before Stage 2, we perform cold-start supervised fine-tuning using responses generated by Qwen3-VL-30B-A3B-Instruct for 2,000 object-pair instances. In Stage 2, we optimize the MLLM via reinforcement learning, where rewards are derived from the relational embedding space learned in Stage 1 (Sec.~\ref{sec:stage2}). Overall, the embedding space learned in Stage 1 serves as a bridge between free-form reasoning and structured relation prediction, and provides guidance for optimizing the generation process in Stage 2.

%Finally, Stage 3 introduces optional supervised fine-tuning to anchor base predicates while preserving transferability to novel relations.

%Our proposed framework formulates open-vocabulary predicate prediction as a discriminability-driven reinforcement learning problem, where an MLLM serves as the relation predictor and an embedding model provides discriminability assessment of the MLLM's outputs. As shown in Fig.~\ref{fig:pipeline}, the framework operates in two stages. In Stage 1, we build a discriminability-aware reward model: we first prompt the MLLM to generate diverse chain-of-thought reasoning and relation predictions for each entity pair, then train the embedding model $f_\phi$ via contrastive learning on these rollouts to establish a semantic space capable of measuring the separability among different relation outputs (Sec.~\ref{sec:stage1}). In Stage 2, we leverage the trained $f_\phi$ as a reward model within a Group Relative Policy Optimization (GRPO) framework to optimize the MLLM, encouraging it to produce outputs that are maximally discriminative across relation categories (Sec.~\ref{sec:stage2}). 
% We further introduce a supervised variant that leverages ground-truth labels of base classes to construct more structured prototypes while maintaining annotation-free generalization to novel classes.

\begin{figure}[t!]
  \centering
  \includegraphics[width=0.9\textwidth]{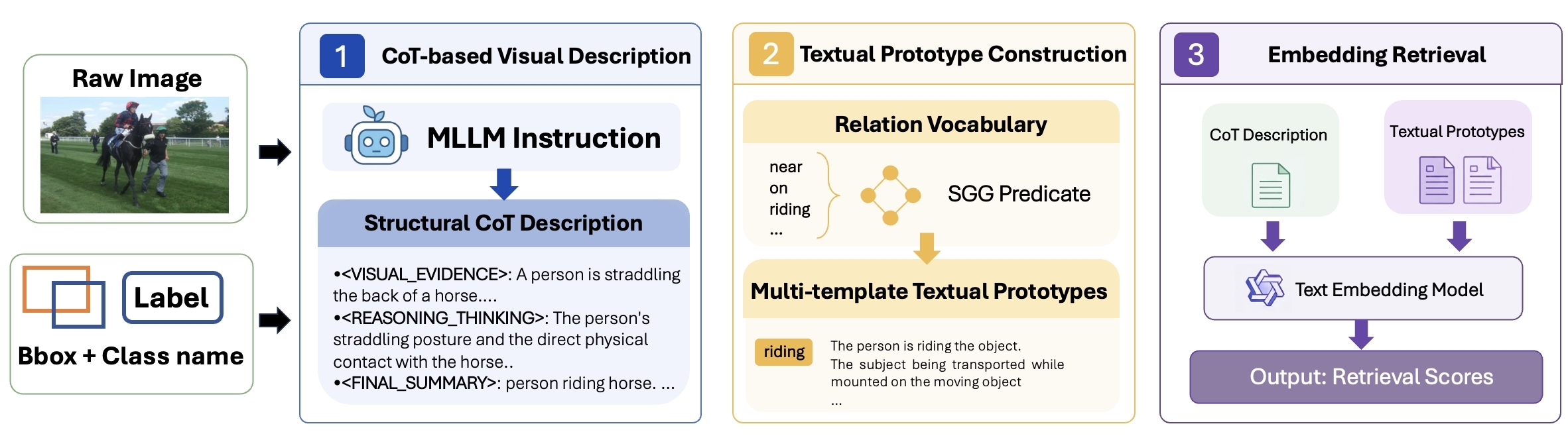}
   \caption{\textbf{Inference pipeline of our method.} (1)CoT-based Visual Description: The MLLM generates a structural CoT Description as the image's textual representation.(2)Textual Prototype Construction: Build up a Multi-template textual prototype for each SGG predicate.(3)Embedding Retrieval: We utilize an embedding model to compute the similarities between the CoT and the text prototypes to yield the final results.}
    \label{fig:inference}
    \vspace{-5mm}
\end{figure}

\subsection{Stage 1: Self-Supervised Relational Space Construction}
\label{sec:stage1}

To support generalization beyond seen predicates, we first aim to learn a transferable relational representation space that captures consistent semantics across diverse instances. A key challenge is that explicit annotations for fine-grained relations are limited in open-vocabulary settings.Our key observation is that the MLLM's own outputs already contain a naturally paired structure---each response couples a chain-of-thought reasoning $\mathbf{y}^\text{cot}$ with a relation prediction $\mathbf{y}^\text{pred}$---providing a self-supervisory signal for contrastive alignment without ground-truth predicate targets. Rather than treating predictions as discrete labels, we utilize this paired structure to learn representations grounded in reasoning, enabling the model to capture instance-level relational semantics while reducing reliance on dataset-specific co-occurrence patterns.

%To correct the MLLM's insufficiently discriminative output distribution, we first need a reward model capable of assessing the separability of its outputs. However, training such a model typically requires fine-grained relation annotations, which are scarce in the open-vocabulary setting.
%Our key observation is that the MLLM's own outputs already contain a naturally paired structure---each response couples a chain-of-thought reasoning $\mathbf{y}^\text{cot}$ with a relation prediction $\mathbf{y}^\text{pred}$---providing a free self-supervisory signal without any external labels. We exploit this structure by first sampling the MLLM to obtain a dataset representing its current output distribution, and then applying contrastive learning over the (reasoning, prediction) pairs to train $f_\phi$ into a discriminative embedding space in a fully Self-supervised manner.

\paragraph{Data Collection.}
We first construct the training data for $f_\phi$ by performing inference with the MLLM over the training set. For each image $I$ and entity pair $(v_i, v_j)$, we query the MLLM with a unified relation-reasoning prompt that specifies the object categories and bounding boxes, and sample a response 
$y = [y^\text{cot}; y^\text{pred}]$. This yields a dataset 
$\mathcal{D} = \{(y^\text{cot}_n, y^\text{pred}_n)\}^{|\mathcal{D}|}_{n=1}$ 
of chain-of-thought reasoning and predicted relation pairs. The full prompt template is provided in Appendix~\ref{appendix:prompt}.

\paragraph{Self-supervised Representation Learning.}
We train $f_\phi$ to align the CoT reasoning $\mathbf{y}^\text{cot}$ with its corresponding relation prediction $\mathbf{y}^\text{pred}$ via contrastive learning. Given a mini-batch of $B$ pairs $\{(\mathbf{y}^\text{cot}_i, \mathbf{y}^\text{pred}_i)\}_{i=1}^{B}$, we compute their embeddings:
\begin{equation}
    \mathbf{e}^\text{cot}_i = f_\phi(\mathbf{y}^\text{cot}_i), \quad \mathbf{e}^\text{pred}_i = f_\phi(\mathbf{y}^\text{pred}_i).
\end{equation}
The training objective is a symmetric InfoNCE loss. Taking the CoT-to-prediction direction as an example:
\begin{equation}
    \mathcal{L}_{\text{c2p}} = -\frac{1}{B} \sum_{i=1}^{B} \log \frac{\exp(\text{sim}(\mathbf{e}^\text{cot}_i, \mathbf{e}^\text{pred}_i) / \tau)}{\sum_{j=1}^{B} \exp(\text{sim}(\mathbf{e}^\text{cot}_i, \mathbf{e}^\text{pred}_j) / \tau)},
\end{equation}
where $\text{sim}(\cdot, \cdot)$ denotes cosine similarity and $\tau$ is a temperature hyperparameter. The prediction-to-CoT direction $\mathcal{L}_{\text{p2c}}$ is defined analogously. The overall objective is:
\begin{equation}
    \mathcal{L}_\text{stage1} = \frac{1}{2}(\mathcal{L}_{\text{c2p}} + \mathcal{L}_{\text{p2c}}).
\end{equation}

This objective aligns each CoT rationale with its corresponding predicted relation text, encouraging
instance-level consistency in the learned representation space. Importantly, by relying on diverse
CoT-prediction pairs rather than directly fitting base-class annotations, the learned space reduces
dependence on annotated predicate frequencies and provides a transferable foundation for downstream
alignment in Stage 2.
%This objective encourages representations where reasoning processes corresponding to the same relational semantics are consistently aligned, while implicitly reducing ambiguity across different relations. Importantly, by relying on self-supervised signals rather than explicit labels, the learned space is less biased toward high-frequency predicates and better captures transferable relational patterns, providing a robust foundation for downstream alignment in Stage 2.

\subsection{Stage 2: Specificity-Driven Policy Optimization}
\label{sec:stage2}

With the transferable representation space established by $f_\phi$, we next aim to align the MLLM’s generation process with this space, such that the produced reasoning reflects instance-specific relational semantics rather than generic or biased patterns. We adopt Group Relative Policy Optimization(GRPO)~\cite{shao2024deepseekmathpushinglimitsmathematical} and design a self-supervised reward grounded in the learned embedding space.

\paragraph{Training Objective.}
We present the training objective and highlight its key difference from vanilla GRPO~\cite{shao2024deepseekmathpushinglimitsmathematical}. Given an image $I$ and a prompt $\mathbf{x}$ specifying an entity pair, the current policy $\pi_\theta$ generates a group of $G$ responses $\{\mathbf{y}_i\}_{i=1}^{G}$, where each $\mathbf{y}_i = [\mathbf{y}_i^\text{cot}; \mathbf{y}_i^\text{pred}]$. A reward $r_i$ is computed for each response, and the group-level advantage is estimated as $\hat{A}_i = ({\boldsymbol{r_i} - \text{mean}(\{\boldsymbol{r_j}\})}) / {\text{std}(\{\boldsymbol{r_j}\})}$. The training objective is:
\begin{equation}
\begin{aligned}
\mathcal{L}_{\mathrm{stage2}}(\theta)
&= -\mathbb{E}\!\left[
\frac{\sum_{i=1}^{G}\sum_{t=1}^{T_i}
\left(\tilde u_{i,t}-\beta\widehat D_{\mathrm{KL},i,t}\right)}
{\sum_{i=1}^{G}T_i}
\right],\\
u_{i,t}
&=\min\!\left(\rho_{i,t}\hat A_i,\,
\operatorname{clip}(\rho_{i,t},0.8,1.3)\hat A_i\right),\\
\tilde u_{i,t}
&=\begin{cases}
\max(u_{i,t},3\hat A_i), & \hat A_i<0,\\
u_{i,t}, & \hat A_i\geq0.
\end{cases}
\end{aligned}
\end{equation}
Here, $T_i$ is the number of valid response tokens and
$\rho_{i,t}=
\pi_\theta(y_{i,t}\mid I,\mathbf{x},\mathbf{y}_{i,<t})/
\pi_{\theta_{\mathrm{old}}}(y_{i,t}\mid I,\mathbf{x},\mathbf{y}_{i,<t})$.
The expectation is over training samples and responses drawn from $\pi_{\theta_{\mathrm{old}}}$.
We average over valid response tokens and use a low-variance KL estimator relative to $\pi_{\mathrm{ref}}$ with $\beta=0.01$.
The response-level advantage $\hat A_i$ is computed from our specificity-based reward.

\paragraph{Representation-Aligned Reward.}
The reward encourages object-pair-specific responses without directly evaluating ground-truth predicate correctness. We compute this reward in the embedding space learned by $f_\phi$ in Stage 1. Concretely, consider a training batch consisting of $M$ samples (entity pairs), each with $G$ rollouts, yielding $M \times G$ responses in total. We encode each complete response $\mathbf{y}_i$ (the concatenation of CoT reasoning and relation prediction) into a normalized embedding:
\begin{equation}
    \mathbf{e}_i = f_\phi(\mathbf{y}_i), \quad i = 1, \ldots, M \times G.
\end{equation}
We sample each batch without repeated annotated subject--predicate--object category triplets to reduce false negatives. For a given rollout $\mathbf{y}_i$ belonging to sample $m$, we define its negative set $\mathcal{N}_i$ as all rollouts from \textit{other} samples in the batch (i.e., rollouts from different entity pairs). The specificity reward is then computed as:
\begin{equation}\label{eq:8}
    r_i^\text{spec} = 1 - \tau \cdot \left( \log \frac{1}{|\mathcal{N}_i|} \sum_{k \in \mathcal{N}_i} \exp\!\big(\text{sim}(\mathbf{e}_i, \mathbf{e}_k) / \tau\big) \right),
\end{equation}
where $\tau$ is a temperature parameter and $\text{sim}(\cdot, \cdot)$ denotes cosine similarity. The inner term is a LogMeanExp aggregation that provides a smooth approximation to the maximum similarity with negative samples. 
This formulation discourages generic responses shared across different entity pairs by assigning lower rewards to responses with higher cross-pair similarity. It encourages object-pair-specific relational descriptions and supports learning more discriminative representations for downstream relation retrieval.
%This formulation encourages the model to produce responses that are object-pair-specific and semantically grounded, avoiding collapse into generic patterns shared across different relational contexts. In practice, responses dominated by high-frequency or ambiguous relational descriptions tend to exhibit higher similarity with other samples and thus receive lower rewards, while those grounded in distinctive visual evidence are favored. As a result, the model is guided to reduce shortcut learning from dataset biases and to generate more transferable and noise-robust CoT representations.

%Intuitively, a response receives a high reward when its embedding is well-separated from responses to other entity pairs, indicating that its output carries distinctive semantic content for the queried relation. Conversely, vague or generic responses that are easily confused with outputs for other relations receive lower rewards.

The specificity reward does not directly verify predicate correctness or visual grounding; a distinctive rationale may still be visually incorrect.
In addition to the Representation-Aligned Reward, we include a format reward $r_i^\text{fmt}$ that checks whether the response follows the prescribed output structure (e.g., producing a valid chain-of-thought followed by a relation prediction). The final reward is:
\begin{equation}\label{eq:9}
    r_i = r_i^\text{spec} + \lambda\, r_i^\text{fmt},
\end{equation}
where $\lambda$ controls the relative weight of format compliance.

\subsection{Relation Prediction}
\label{sec:inference}

%At test time, we construct relation category prototypes for the full predicate vocabulary $\mathcal{C}^p = \mathcal{C}^p_\text{base} \cup \mathcal{C}^p_\text{novel}$ by encoding each relation name through the embedding model: $\mathbf{c}_k = f_\phi(\text{name}_k)$ for $k = 1, \ldots, |\mathcal{C}^p|$. Given an image $I$ and an entity pair $(v_i, v_j)$, the optimized MLLM generates a single response $\mathbf{y} = [\mathbf{y}^\text{cot}; \mathbf{y}^\text{pred}]$ via greedy decoding. The complete response is then mapped to the embedding space as $\mathbf{e} = f_\phi(\mathbf{y})$, and the predicate is determined by the highest-scoring category:
%\begin{equation}
%    p_{ij} = \arg\max_{k} \; \text{sim}(\mathbf{e}, \mathbf{c}_k).
%\end{equation}
%This procedure is applied to all candidate entity pairs to produce the final scene graph.

At test time, we construct predicate prototypes for the full predicate vocabulary 
$\mathcal{C}^p = \mathcal{C}^p_\text{base} \cup \mathcal{C}^p_\text{novel}$ 
using multiple textual templates for each predicate. Specifically, for each relation category $k$, 
we encode a set of template descriptions $\{\mathbf{c}_{k,m}\}_{m=1}^{M}$ with the embedding model, 
where $\mathbf{c}_{k,m}=f_\phi(T_m(\text{name}_k))$. Given an image $I$ and an entity pair 
$(v_i, v_j)$, the optimized MLLM generates a single response 
$\mathbf{y} = [\mathbf{y}^\text{cot}; \mathbf{y}^\text{pred}]$ via greedy decoding. 
The complete response is mapped to the embedding space as $\mathbf{e}=f_\phi(\mathbf{y})$. 
For each predicate, its score is computed as the maximum similarity between the response embedding 
and all its template prototypes, and the final predicate is determined by:
\begin{equation}
    p_{ij} = \arg\max_{k} \; \max_{m=1,\ldots,M} \text{sim}(\mathbf{e}, \mathbf{c}_{k,m}).
\end{equation}
This procedure is applied to all candidate entity pairs to produce the final scene graph.We further apply pair-level score normalization and a lightweight subject-object prior calibration during inference, with details provided in Appendix~\ref{appendix:Additional Implementation Details}.

\subsection{Supervised Variant}
To align with the standard OV-SGG evaluation protocol, where base class annotations are available during training, we introduce a supervised extension. Starting from the embedding model $f_\phi$ trained in Stage 1, we utilize the semantic embeddings of the base class relation names as textual anchors, denoted as $\{\mathbf{a}_k\}_{k=1}^{|\mathcal{C}^p_\text{base}|}$, where each anchor is generated by encoding the predicate text through $f_\phi$. We then fine-tune $f_\phi$ with an InfoNCE objective analogous to 
(Eq.~(5)), aligning each MLLM-generated reasoning output embedding with its corresponding ground-truth relation name embedding while contrasting it against all other base class candidates. At inference time, the model serves as a relation extractor, determining the predicate by identifying the nearest neighbor in the embedding space. Since novel class embeddings are constructed identically from their respective relation names, this approach naturally extends the model's discriminative power to unseen relations without requiring any novel class annotations.

%% file: sections/exps.tex
\section{Experiments}

\subsection{Experiment Setup}
\paragraph{Datasets}

We evaluate our method on Visual Genome (VG150)~\cite{krishna2017visual} and PSG~\cite{yang2022panoptic}.
For VG150, we follow the standard \textit{OVR-SGG} protocol~\cite{chen2024expanding}, we use the EPIC~\cite{yu2023visuallyEPIC} split for PredCls and the OvSGTR~\cite{chen2024expanding} split for SGDet. Both contain 35 base and 15 novel predicates, with different memberships (Appendix C.1). During training, only base-predicate annotations are available, while evaluation is conducted on both base and novel predicates to measure open-vocabulary generalization. We report results under two standard settings: Predicate Classification (PredCls) and Scene Graph Detection (SGDet)~\cite{xu2017scene}. For PSG, we follow PGSG's predicate split~\cite{li2024pixelsgraphsopenvocabularyscene} and evaluate under SGDet. In addition, we provide supplementary Human-Object Interaction (HOI) experiments on HICO-DET~\cite{chao2018learning} in Appendix \ref{appendix:zs_hoi} to further examine the transferability of our framework across visual relationship understanding tasks.

\paragraph{Evaluation metric}
We adopt mean Recall@K ($mR@K$) and Recall@K ($R@K$) as the primary evaluation metrics, where $K \in \{50, 100\}$. 
While $R@K$ measures overall triplet-level recall and is often dominated by frequent predicates, $mR@K$ computes recall independently for each predicate category and then averages over all classes. 
Therefore, $mR@K$ provides a more balanced evaluation of low-frequency, long-tailed, and unseen relations, which is particularly aligned with the objective of OV-SGG. 
We thus regard $mR@K$ as the main metric and report $R@K$ as a complementary measure for comprehensive comparison.

\paragraph{Implementation Details}
We adopt Qwen3-VL-4B-Instruct~\cite{bai2025qwen3} as the backbone MLLM and Qwen3-VL-Embedding-2B~\cite{li2026qwen3} as the embedding model $f_\phi$. The embedding model is fine-tuned with LoRA on 2 NVIDIA L20 GPUs. The MLLM is fully fine-tuned with GRPO on 4 NVIDIA L40 GPUs. For SGDet, we use the object detector from ACC~\cite{li2025interaction} to provide object proposals, isolating improvements in predicate prediction. For VG150 SGDet, we use TP, PR, SSL, and RL without supervised embedding tuning (ST). More implementation details are provided in Appendix \ref{appendix:Additional Implementation Details}.

%We adopt Qwen3-VL-4B-Instruct\cite{bai2025qwen3} as the backbone MLLM and Qwen3-VL-Embedding-2B\cite{li2026qwen3} as the embedding model $f_{\phi}$, which inherently supports text-pair similarity matching. For training, $f_{\phi}$ is fine-tuned via LoRA with rank $r=8$, $\alpha=16$, and dropout $0.1$. We adopt in-batch negatives as contrastive negative samples with temperature $\tau=0.1$. MLLM undergoes full-parameter fine-tuning via GRPO. All models are trained on 4 NVIDIA L40 GPUs.
%We implement the GRPO training based on the EasyR1\cite{zheng2025easyr1} reinforcement learning training framework, the LogMeanExp aggregation for the discriminability reward (Eq.~\ref{eq:8}) is set to $0.1$ and in the reward computation (Eq.~\ref{eq:9}), the discriminability reward is assigned a primary weight of $\lambda_{div}=1.0$, while the format compliance reward weight $\lambda$ (or $\lambda_{fmt}$) is set to $0.3$ to maintain structural generation stability. Furthermore, we adapt the object detector from the previous work ACC \cite{li2025interaction} to provide object proposals, ensuring a fair comparison by isolating our improvements in predicate discrimination. To further refine predictions during inference, we apply a spatial heuristic to filter out non-overlapping object pairs exceeding a predefined distance threshold. Additionally, we integrate a statistical prior calibration, adding auxiliary bonus scores to the final embedding similarities based on subject-predicate-object co-occurrence probabilities derived from the training set. More details are displayed is appendix.

\subsection{Comparisons with State-of-the-Art OV-SGG Methods}

\begin{table*}[t!]
\centering
\caption{\textbf{OVR-SGG results under SGDet on the full VG150 test set.}
All results use graph-constraint evaluation. $\dagger$ denotes our re-evaluation of ACC. ACC+RaRe uses the same detected boxes, object labels, and candidate pairs as ACC.}\resizebox{\linewidth}{!}{
\begin{tabular}{@{}l l c c c c c@{}}
\toprule
{\textbf{Model}} & \multicolumn{2}{c}{\textbf{Total (Relation)}} & \multicolumn{2}{c}{\textbf{Base (Relation)}} & \multicolumn{2}{c}{\textbf{Novel (Relation)}} \\
\cmidrule(lr){2-3} \cmidrule(lr){4-5} \cmidrule(l){6-7}
&  R@50/100 & mR@50/100 & R@50/100 & mR@50/100 & R@50/100 & mR@50/100 \\

\midrule
OvSGTR\cite{chen2024expanding} & 20.46/23.86 & 3.91/4.62 & 26.14/30.16 & 4.81/5.60 & 13.45/16.19 & 1.82/2.32 \\
OvSGTR+RAHP\cite{liu2025relationawarehierarchicalpromptopenvocabulary} & 21.50/25.74 & 4.51/5.37 & 26.29/30.16 & 5.15/5.94 & 15.59/19.92 & 3.01/4.04 \\
OwSGG(Qwen72b)\cite{dutta2025open} & -/- & -/- & -/- & -/- & 2.19/3.06 & 2.19/3.07 \\  
OpenSGen\cite{openSGen} & 18.0/20.5 & 6.2/7.7 & -/- & -/- & 15.7/17.9 & -/- \\
ACC(Swin-B)\cite{li2025interaction}$^\dagger$ & \textbf{24.04}/\textbf{28.50} & 4.30/5.11 & 27.41/31.53 & 5.08/5.95 & \textbf{19.15}/\textbf{23.72} & 2.48/3.17 \\
ACC+Ours & 19.57/22.96 & \textbf{10.38}/\textbf{12.32} & \textbf{28.81}/\textbf{33.10} & \textbf{11.96}/\textbf{14.17} & 9.94/12.38 & \textbf{6.69}/\textbf{8.02} \\

%$^\dagger$
 
\bottomrule
\end{tabular}
}
\label{tab:sgdet_results}
\vspace{-15pt}

\end{table*}

\begin{table*}[t]
\centering
\caption{\textbf{ Experimental results of OVR-SGG on PredCls setting.}}
\resizebox{\linewidth}{!}{
\begin{tabular}{@{}l l c c c c c@{}}
\toprule
{\textbf{Model}} & \multicolumn{2}{c}{\textbf{Total (Relation)}} & \multicolumn{2}{c}{\textbf{Base (Relation)}} & \multicolumn{2}{c}{\textbf{Novel (Relation)}} \\
\cmidrule(lr){2-3} \cmidrule(lr){4-5} \cmidrule(l){6-7}
&  R@50/100 & mR@50/100 & R@50/100 & mR@50/100 & R@50/100 & mR@50/100 \\
\midrule
 PGSG\cite{li2024pixelsgraphsopenvocabularyscene} & 26.90/33.90 & 10.80/13.90 & - & - & - & 5.20/7.70 \\
SGTR+\cite{li2024sgtrendtoendscenegraph} & 36.32/41.51 & 13.30/17.60 & 40.50/46.67 & 19.57/24.96 & 0.00/0.00 & 0.00/0.00 \\
SGTR+ + RAHP\cite{liu2025relationawarehierarchicalpromptopenvocabulary} & 39.92/46.03 & 16.88/22.18 & 41.29/47.65 & 20.51/25.18 & 15.46/20.37 & 11.82/15.46 \\
SVPR\cite{he2022towards} & 33.50/35.90 & 8.30/10.80 & -/- & -/- & -/- & -/- \\
Epic\cite{yu2023visuallyEPIC} & -/- & 16.50/21.80 & 28.30/31.10 & -/- & 13.90/18.30 & -/- \\
PE-NET\cite{zheng2023prototypebasedembeddingnetworkscene} & 58.79/61.23 & 19.18/20.97 & 63.62/67.09 & 23.18/25.79 & 0.00/0.00 & 0.00/0.00 \\
PE-NET+RAHP\cite{liu2025relationawarehierarchicalpromptopenvocabulary} & \textbf{64.70}/\textbf{69.11} & 24.50/\textbf{28.25} & \textbf{65.15}/\textbf{70.54} & 24.99/\textbf{30.19} & \textbf{20.79}/\textbf{29.00} & 15.70/23.73 \\
Ours & 49.90/51.33 & \textbf{24.62}/25.80 & 50.36/51.81 & \textbf{25.53}/26.45 & 12.75/13.60 & \textbf{22.49}/\textbf{24.27} \\
%$^\dagger$/ 
\bottomrule
\end{tabular}
}
\label{tab:predcls_results}
\vspace{-15pt}

\end{table*}

Comprehensive comparison between our framework and state-of-the-art (SOTA) OV-SGG methods under the standard SGDet and PredCls setting are in Table\ref{tab:sgdet_results} and Table \ref{tab:predcls_results}. The key findings and conclusions are summarized as follows:

\paragraph{Performance under Detection Settings}
As shown in Table~\ref{tab:sgdet_results}, RaRe achieves 14.17\% Base mR@100 and 8.02\% Novel mR@100 on VG150 using detector-generated proposals.
Under fixed ACC proposals, training-free MLLM retrieval already accounts for most of the Total mR gain over ACC; adding SSL, RL, and PR yields modest further gains while reducing Novel mR (Table~\ref{tab:vg_fixed_proposal}).
In contrast, the same-proposal PSG ablation shows improvements in both Total and Novel mR over training-free retrieval (Table~\ref{tab:psg_ablation}), indicating that the benefits of learning and calibration vary across datasets and proposal settings.

\paragraph{Improvement on Novel Class Generalization}
We further examine novel predicate generalization under both SGDet and PredCls settings, as shown in Tables~\ref{tab:sgdet_results} and~\ref{tab:predcls_results}. Under SGDet on VG150, RaRe achieves 10.38/12.32 Total mR@50/100, with 14.17\% Base mR@100 and 8.02\% Novel mR@100, improving category-balanced recognition of unseen predicates in the end-to-end setting. 

Under PredCls, where ground-truth object boxes are provided, closed-set SGG methods adapted to the open-vocabulary setting, such as PE-NET~\cite{zheng2023prototypebasedembeddingnetworkscene} and SGTR+~\cite{li2024sgtrendtoendscenegraph}, show limited generalization to novel predicates. Compared with the strongest VLM-based baseline PE-NET+RAHP~\cite{liu2025relationawarehierarchicalpromptopenvocabulary}, which achieves 15.70\% Novel mR@50, RaRe improves this metric to 22.49\%, yielding an absolute gain of 6.79 percentage points. These results indicate that reasoning-aware relational representations help transfer relational semantics beyond base predicates, leading to stronger generalization to unseen relations.
%Table\ref{tab:predcls_results} shows traditional closed-set SGG methods adapted to the open-vocabulary setting (e.g., PE-NET \cite{zheng2023prototypebasedembeddingnetworkscene}, SGTR \cite{li2024sgtrendtoendscenegraph}) suffer from catastrophic performance degradation on novel categories, with their Novel mR@50 dropping to 0.00\%, as they severely overfit the co-occurrence patterns of base classes seen during training. Even the strongest VLM-based OV-SGG baseline, PE-NET+RAHP \cite{liu2025relationawarehierarchicalpromptopenvocabulary}, only achieves 15.70\% on Novel mR@50. In contrast, our proposed discriminability-driven framework elevates this metric to 22.49\%, yielding a substantial absolute improvement of +6.79\% over the strongest baseline. This result validates that our method guides the MLLM to learn generalizable semantic evidence of relational interactions, rather than simply memorizing high-frequency co-occurrence shortcuts from the training data.

\paragraph{Balanced Recognition across Long-Tailed Distributions}
As shown in Table~\ref{tab:predcls_results}, RaRe achieves stronger category-balanced performance despite a lower Base R@50 than PE-NET+RAHP, reaching 25.53\% Base mR@50 and 24.62\% Total mR@50. 
This indicates that our method improves recognition beyond dominant head predicates and better covers the long-tailed predicate distribution. The contrast between lower global R@K and higher mR@K suggests that RaRe shifts the prediction behavior from fitting frequent co-occurrence patterns toward more balanced relational generalization.
%In Table\ref{tab:predcls_results}, we observe that while our Base R@50 (50.36\%) shows a decline compared to PE-NET+RAHP (65.15\%), it is critical to clarify the intrinsic difference between the two core metrics. Specifically, R@K is a global recall metric that is inherently biased towards high-frequency base classes that dominate the training distribution, and thus cannot reflect the model's balanced recognition capability across all categories. In contrast, mR@K assigns equal weight to all predicate categories, which is the core evaluation metric for OV-SGG tasks, as it faithfully reflects the model's performance on long-tailed, low-frequency, and unseen novel relations. our method achieves 25.53\% on Base mR@50 and 24.62\% on Total mR@50, both outperforming existing methods on category-balanced mR. This phenomenon of "decreasing global R@K with increasing category-averaged mR@K" demonstrates that our method effectively mitigates the prevalent "seesaw effect" in OV-SGG, where models sacrifice base-class performance for novel-class generalization. By suppressing the model's shortcut learning on high-frequency head classes, we achieve truly unbiased and balanced recognition across the full long-tailed relation distribution.

\paragraph{Results on PSG}
We evaluate RaRe under SGDet on PSG using PGSG's predicate split, retaining its detected boxes, object labels, and candidate pairs. As shown in Table~\ref{tab:psg_results}, PGSG+RaRe achieves the best reported Total mR, Novel mR, and Novel R among the compared methods, although its Total R remains lower than PGSG.These results extend the gains in category-balanced and novel-predicate recognition to another dataset and upstream detector.

\begin{table}[t]
\centering
\caption{\textbf{OVR-SGG results under SGDet on PSG.}
PGSG+RaRe uses the same object proposals and candidate pairs as PGSG.}
\label{tab:psg_results}
\small
\setlength{\tabcolsep}{5pt}
\begin{tabular}{@{}lcccc@{}}
\toprule
\textbf{Model}
& \multicolumn{2}{c}{\textbf{Total (Relation)}}
& \multicolumn{2}{c}{\textbf{Novel (Relation)}} \\
\cmidrule(lr){2-3}\cmidrule(lr){4-5}
& R@50/100 & mR@50/100 & R@50/100 & mR@50/100 \\
\midrule
SGTR+~\cite{li2024sgtrendtoendscenegraph}
& 14.20/18.20 & 6.40/8.40 & --/-- & 0.00/0.00 \\
PGSG~\cite{li2024pixelsgraphsopenvocabularyscene}
& 18.00/20.20 & 13.50/16.40 & --/-- & 7.40/11.30 \\
OvSGTR~\cite{chen2024expanding}
& 17.76/19.50 & --/-- & 6.93/8.08 & --/-- \\
ACC~\cite{li2025interaction}
& \textbf{20.01/21.71} & --/-- & 8.78/9.70 & --/-- \\
PGSG+RaRe
& 16.34/18.65 & \textbf{17.52/20.59}
& \textbf{9.11/10.34} & \textbf{14.76/16.10} \\
\bottomrule
\end{tabular}
\end{table}

\subsection{Ablation Study}

We conduct ablation studies under the PredCls setting to validate each component of our framework, with results summarized in Table~ \ref{tab:ablation_results}.

\begin{wrapfigure}{r}{0.5\textwidth}
  \vspace{-15pt} % 可选：如果图片上方空白太大，可以用这个微调上移
  \centering
  \includegraphics[width=\linewidth]{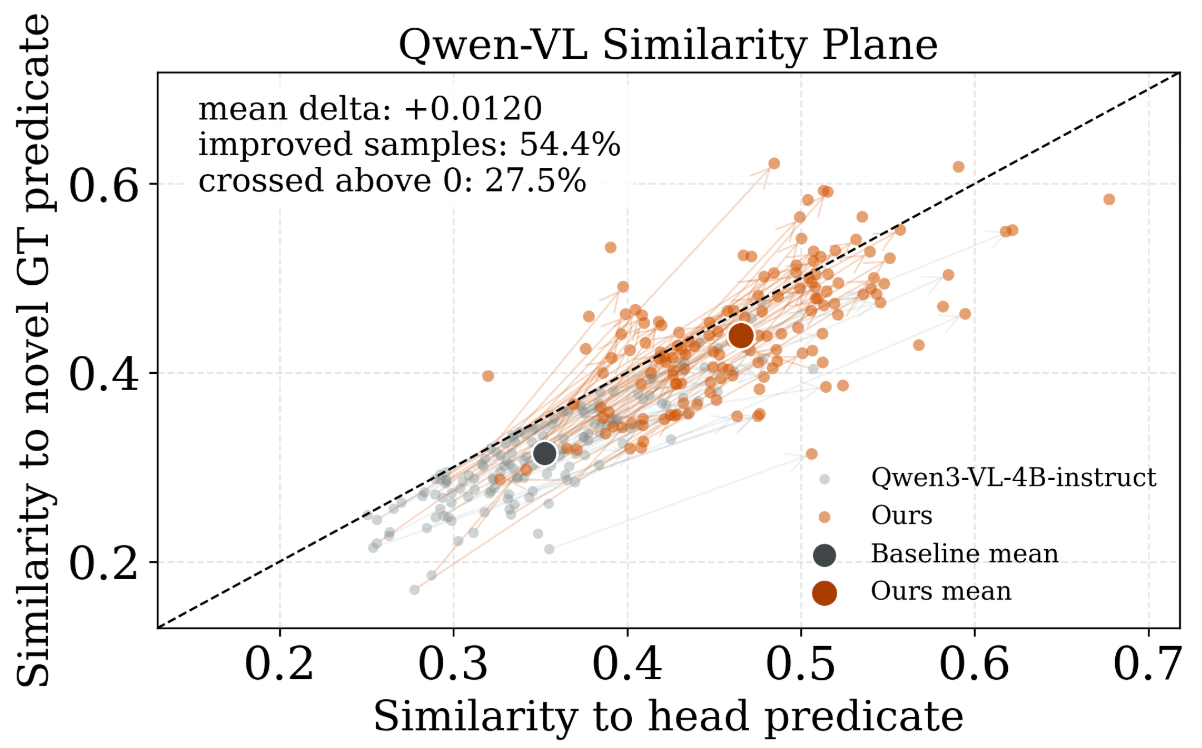} % 这里的\linewidth是wrapfigure内部的宽度
  \caption{Distribution of CoT samples in the similarity plane.}
  \label{fig:cot_escape}
  \vspace{-10pt} % 可选：微调图片下方的空白
\end{wrapfigure}

\paragraph{Effectiveness of Core Modules}
Compared with TP-only retrieval, RaRe-Core(row 7) improves Total and Novel mR@50/100 by 3.90/3.93 and 5.01/4.41 points, respectively, demonstrating the contribution of reasoning representation learning without PR or ST. Adding PR and ST further improves Base mR@50/100 from 21.81/23.14 to 25.53/26.45, while reducing Novel mR@50/100 from 27.11/28.46 to 22.49/24.27. These results distinguish the gains from the core learning stages from the base--novel trade-off introduced by downstream tuning and calibration. We report the full configuration as a compromise between base and novel recognition: it improves Base mR over RaRe-Core while retaining higher Novel mR than the SSL+ST variant.

\begin{table*}[t]
\centering
\caption{\textbf{\textbf{Ablation study of RaRe under PredCls on the full VG150 test set.}} We evaluate the impact of different components: TP (predicate template for multi-template prototype construction), PR (prior calibration), SSL (self-supervised embedding learning in Stage 1), RL (GRPO in Stage 2), and ST (supervised embedding tuning). We report category-balanced mR@50/100 in the main paper.}
\label{tab:ablation_results}
\small
\setlength{\tabcolsep}{4pt}
\renewcommand{\arraystretch}{1.05}
\begin{tabular}{@{} c c c c c c c c c @{}}
\toprule
\textbf{\#} & \textbf{TP} & \textbf{PR} & \textbf{RL} & \textbf{SSL} & \textbf{ST}
& \textbf{Total mR@50/100}
& \textbf{Base mR@50/100}
& \textbf{Novel mR@50/100} \\
\midrule
1 & & & & & & 19.42/20.25 & 17.94/19.02 & 22.88/23.13 \\
2 & \checkmark & & & & & 19.50/20.80 & 18.38/19.41 & 22.10/24.05 \\
3 & \checkmark & \checkmark & & & & 20.68/23.54 & 23.37/25.93 & 14.39/17.95 \\
4 & \checkmark & \checkmark & \checkmark & & & 20.80/23.25 & 21.39/23.61 & 19.41/22.41 \\
5 & \checkmark & \checkmark & & \checkmark & & 21.54/24.23 & 23.61/26.08 & 16.72/19.92 \\
6 & \checkmark & \checkmark & & \checkmark & \checkmark & \textbf{27.29/29.14} & \textbf{33.11/34.61} & 13.71/16.38 \\
7 & \checkmark & & \checkmark & \checkmark & & 23.40/24.73 & 21.81/23.14 & \textbf{27.11/28.46} \\
8 & \checkmark & \checkmark & \checkmark & \checkmark & \checkmark & 24.62/25.80 & 25.53/26.45 & \textbf22.49/24.27 \\
\bottomrule
\end{tabular}
\vspace{-10pt}
\end{table*}

\paragraph{Head Predicate Prediction Shift Analysis}
As shown in Figure~\ref{fig:cot_escape}, we project novel ground-truth (GT) samples onto a 2D similarity plane defined by $\mathrm{sim}(\mathrm{CoT}, \mathrm{head})$ and $\mathrm{sim}(\mathrm{CoT}, \mathrm{novel\ GT})$. The diagonal dashed line denotes equal similarity, so points above this line indicate CoT representations that are semantically closer to the novel predicate than to the head predicate. Compared with the baseline, our method exhibits a systematic shift toward higher similarity to the novel GT, together with a more dispersed distribution. This observation suggests that RL-aligned CoT representations reduce relative reliance on head predicates and better capture novel relational semantics.

\paragraph{Same-Proposal Ablation on PSG}
With PGSG proposals and TP/PR fixed, full RaRe improves Total mR@50/100 from 12.56/14.39 to 17.52/20.59 over training-free retrieval (Table~\ref{tab:psg_ablation}).
Adding RL to SSL+ST improves Novel mR@50/100 by 5.29/5.61 points, mitigating the decline introduced by supervised tuning.
Full RaRe achieves the highest Total mR, while omitting ST yields slightly higher Novel mR@100.

\begin{table}[t]
\centering
\caption{\textbf{Same-proposal SGDet ablation on PSG.}
All variants use identical PGSG proposals and evaluation protocols, with TP and PR fixed.
SSL, RL, and ST denote self-supervised embedding learning, policy optimization, and supervised embedding tuning, respectively.}
\label{tab:psg_ablation}
\small
\setlength{\tabcolsep}{6pt}
\renewcommand{\arraystretch}{1.05}
\begin{tabular}{@{}cccccc@{}}
\toprule
\textbf{\#} & \textbf{RL} & \textbf{SSL} & \textbf{ST}
& \textbf{Total mR@50/100}
& \textbf{Novel mR@50/100} \\
\midrule
1 & & & & 12.56/14.39 & 12.69/14.20 \\
2 & & \checkmark & & 14.02/16.68 & 14.25/15.88 \\
3 & & \checkmark & \checkmark & 15.12/17.04 & 9.47/10.49 \\
4 & \checkmark & & & 12.63/14.45 & 12.83/14.31 \\
5 & \checkmark & \checkmark & & 13.89/15.92 & 14.75/\textbf{16.34} \\
6 & \checkmark & \checkmark & \checkmark
& \textbf{17.52/20.59} & \textbf{14.76}/16.10 \\
\bottomrule
\end{tabular}
\end{table}

%% file: sections/conclusion.tex
% 根据Table2的结果写，分析的结论，强基线弱基线现象
\vspace{-1mm}
\section{Conclusion}
\vspace{-1mm}

In this work, we present RaRe, a reasoning-aware relational representation learning framework for open-vocabulary scene graph generation. Rather than treating MLLM outputs merely as final textual predictions, RaRe transforms chain-of-thought rationales into structured relation embeddings that capture object-pair-specific relational evidence. By aligning CoT rationales with relation semantics through self-supervised contrastive learning and further refining rationale generation with representation-guided policy optimization, our framework converts free-form MLLM reasoning into discriminative, retrieval-compatible relation representations. Experiments on VG150 and PSG demonstrate improved category-balanced predicate recognition, with ablations showing that the additional benefits of learning and calibration depend on the evaluation setting. These results suggest that rationale-derived relational representations provide an effective bridge between generative visual reasoning and structured open-vocabulary relation prediction.

% In this work, we present a CoT-centered relational representation framework for open-vocabulary scene graph generation. Instead of treating MLLM outputs merely as final textual predictions, our method leverages chain-of-thought rationales as transferable semantic representations that capture object-pair-specific relational evidence. Through self-supervised CoT-relation alignment, representation-aligned policy optimization, and supervised retrieval tuning under the standard OV-SGG protocol, our framework converts free-form MLLM reasoning into retrieval-compatible relation representations. Experiments on VG150 demonstrate that our method improves category-balanced recognition on both base and novel predicates, while maintaining strong robustness under the challenging SGDet setting. These results suggest that CoT-based relational representations provide an effective bridge between generative visual reasoning and structured open-vocabulary relation prediction.

\vspace{-1mm}
\paragraph{Limitation}
\vspace{-1mm}
This work has two main limitations. First, although our framework improves the utility of MLLM-generated rationales for relation prediction, it does not explicitly address all sources of visual hallucination in MLLMs, which may still cause fine-grained inaccuracies for small or occluded objects. Second, balancing category-level generalization and overall recall remains challenging, and the benefits of learning and calibration vary across proposal settings. Under fixed ACC proposals on VG150, novel-predicate recognition remains below the training-free retrieval baseline (Appendix~\ref{appendix:additional exp}), possibly reflecting the mismatch between ground-truth training pairs and detector-generated test pairs. Future work could explore adaptive calibration and better adaptation to noisy proposals. The framework may benefit image retrieval and assistive scene understanding, but incorrect or biased predictions can propagate to downstream applications.

% This work has two main limitations. First, it does not explicitly mitigate hallucination in MLLMs, which may lead to fine-grained inaccuracies for small or occluded objects. Second, while this design improves novel-class generalization and long-tailed recognition, it may not fully exploit the statistical advantages of frequent predicates, resulting in a gap in global R@K compared to VLM-based methods. A promising direction is to incorporate frequency-aware calibration more adaptively while preserving the transferability and robustness of CoT-based relational representations.

%% file: sections/acknowledge.tex
\section{Acknowledgments}
This work was supported by the Shanghai Key Technology R\&D Program No.25JC3200500, and the MoE Key Lab of Intelligent Perception and Human-Machine Collaboration (ShanghaiTech University).

%% file: sections/appendix.tex
\newpage
\section{Overview of Appendix}
In this supplementary document, we provide additional details and experimental results to enhance understanding and
insights into our proposed RaRe. This supplementary document is organized as follows:
\begin{itemize}[leftmargin=5mm]

    \item \ref{appendix:prompt}: Instruct Prompt for Inference 

    \item \ref{appendix:Additional Implementation Details}: Additional Implementation Details

    \item \ref{appendix:mllm_backbone}: Zero-Shot PredCls Performance of Different MLLM Backbones

    \item \ref{appendix:Sensitivity analysis}: Sensitivity analysis

    \item \ref{appendix:sftvsgrpo}: Comparison with Alternative MLLM Optimization Strategies

    \item \ref{appendix:sgcls}: SGCls Setting Results

    \item \ref{appendix:zs_hoi}: Cross-Task Transfer: Training-Free Zero-Shot HOI

    \item \ref{appendix:Additional Analysis on Representation Properties}: Additional Analysis on Representation Properties

    \item \ref{appendix:additional exp}: Additional Controls and Inference Cost

    \item \ref{appendix:acccompare}: Novel-Predicate Coverage under SGDet

    \item \ref{appendix:compare}: More Qualitative Comparison Results
\end{itemize}

\section{Instruct Prompt for Inference}
\label{appendix:prompt}
This section shows the unified instruct prompt we use during the inference stage, which is designed to guide the MLLM to generate relation descriptions with chain-of-thought (CoT) reasoning based on the input entity pairs.

\begin{Verbatim}[breaklines=true, frame=single, fontsize=\small]
In this image, there are two objects:
- <|object_ref_start|>{subject_class}<|object_ref_end|> at <|box_start|>{subject_bbox}<|box_end|>
- <|object_ref_start|>{object_class}<|object_ref_end|> at <|box_start|>{object_bbox}<|box_end|>

Based on the image, please analyze the predicate relationship.
IMPORTANT: You must strictly follow the format below using the specific headers enclosed in brackets. Do not output any conversational filler.

<VISUAL_EVIDENCE>
Describe the spatial layout and visual details. Focus primarily on the two objects mentioned above.</VISUAL_EVIDENCE>
<REASONING_THINKING>
Provide step-by-step logical reasoning. Connect the visual evidence to the most likely relationship. Explain the interaction mechanism. Fully consider the relationships between objects.</REASONING_THINKING>
<FINAL_SUMMARY>
State the final relationship predicate in a 'subject predicate object' format and a brief summary sentence describing the interaction.</FINAL_SUMMARY>
\end{Verbatim}

\section{Additional Implementation Details}
\label{appendix:Additional Implementation Details}

\subsection{Predicate Splits.}
For VG150, both settings divide the 50 predicates into 35 base and 15 novel categories, but use different splits.
Under PredCls, we follow EPIC~\cite{yu2023visuallyEPIC} and use \emph{across, along, and, eating, flying in, growing on, laying on, lying on, mounted on, on back of, painted on, part of, says, using, walking in} as novel predicates.
Under SGDet, we follow OvSGTR~\cite{chen2024expanding} and use \emph{belonging to, part of, riding, walking in, in, of, painted on, playing, for, walking on, says, attached to, eating, on, wears} as novel predicates.
The remaining 35 predicates form the base vocabulary in each setting.
Training, prior construction, and evaluation use the corresponding split.

\subsection{Self-supervised Training on Embedding}
\label{sec:c1_emb_unsupervised}
To provide a fine-grained and semantically-aware reward signal for Group Relative Policy Optimization (GRPO), we first train a specialized embedding model via self-supervised contrastive learning. 
This model serves as the reward encoder for evaluating the quality of generated Chain-of-Thought (CoT) reasoning in the subsequent GRPO stage.

\textbf{Training Objective and Data.}
We construct training pairs using only base-set samples, where CoT reasoning is generated by a pretrained Qwen3-VL-4B-Instruct model~\cite{bai2025qwen3} (official release).
Given an image and an object pair, the model produces a structured CoT as shown in \ref{appendix:prompt}, the <VISUAL\_EVIDENCE> and <REASONING\_THINKING> form the \textit{query} $r_i$, while the <FINAL\_SUMMARY> is the corresponding \textit{label} $s_i$. 
The objective is to align the latent representations of reasoning texts with their corresponding relational semantics. 
The training set consists of about 30,000 images, including approximately 46,000 base-set triplet samples, following the original data distribution.
To alleviate predicate long-tail bias, we perform image-level soft-balanced sampling by assigning each image a weight based on the inverse frequencies of its ground-truth predicates, so that images containing more rare predicates are sampled more frequently, and resample with replacement at a fixed dataset size.

\textbf{Model and Optimization.}
The embedding encoder is initialized from the pretrained Qwen3-VL-Embedding-2B model~\cite{li2026qwen3} (official release) and fine-tuned using LoRA, where the backbone parameters are frozen and low-rank adapters are applied to attention projection layers. 
We set the LoRA rank to 8, the scaling factor to 16, and the dropout rate to 0.1. 
Text representations are obtained by selecting the last valid token (according to the attention mask), followed by $\ell_2$ normalization along the embedding dimension.
We train the model using AdamW with a learning rate of $2\times10^{-4}$, a cosine learning rate schedule, and warmup. 
Unless otherwise specified, we use a per-device batch size of 16, gradient accumulation of 2, and train for 3 epochs. 
Gradient clipping is applied with a maximum norm of 1.0.

\textbf{Contrastive Learning.}
We optimize the embedding model using an InfoNCE-style contrastive objective with temperature $\tau = 0.1$. 
For a mini-batch of size $N$, each aligned pair $(r_i, s_i)$ forms a positive pair, while all other labels in the batch serve as in-batch negatives. 
Similarity is computed via temperature-scaled dot product. 

\textbf{Connection to GRPO.}
After training, the embedding model is frozen and used as the encoder in the reward model for GRPO. 
Given a generated CoT, the encoder maps it into the embedding space, where its alignment with relational semantics is evaluated to compute the proposed representation-aligned specificity reward. 
This enables GRPO to optimize CoT generation toward semantically discriminative and relation-specific outputs.

\subsection{GRPO for CoT}
For GRPO training, we first randomly sample 2000 object-pair instances and use Qwen3-VL-30B-A3B-Instruct\cite{bai2025qwen3} to generate responses for supervised fine-tuning as a cold-start stage. 
We then sample approximately 20,000 object-pair instances for policy optimization. 
During GRPO, each object-pair prompt is decoded into a group of 4 responses with sampling temperature 0.7, and the group-relative advantage is computed from the proposed specificity reward. 
We implement GRPO based on the EasyR1 framework. 
The embedding reward model is kept frozen throughout policy optimization, and no novel predicate annotations are used during this stage. We train for one epoch with a learning rate of $10^{-6}$, weight decay of $10^{-2}$, no learning-rate warmup, and a gradient clipping norm of $1.0$. We use PPO clipping bounds of $[0.8,1.3]$, dual clipping with coefficient $3.0$, and a low-variance KL penalty with weight $0.01$. The specificity-reward temperature is $0.1$, distinct from the rollout temperature of $0.7$.

The format reward has weight $0.3$ and equals one only when the response contains all three complete tag pairs: \texttt{<VISUAL\_EVIDENCE>...</VISUAL\_EVIDENCE>}, \texttt{<REASONING\_THINKING>...</REASONING\_THINKING>}, and \texttt{<FINAL\_SUMMARY>...} \\ \texttt{</FINAL\_SUMMARY>}. Matching is case-insensitive and allows multiline content. Otherwise, both the format and specificity rewards are set to zero.

\subsection{Additional Supervised Training on Embedding}
To mitigate potential degradation in base-class performance after self-supervised training (e.g., reduced base mean recall), an additional supervised refinement stage is introduced.

\textbf{Usage and Scope.}
The embedding model obtained from the self-supervised stage is frozen and used as the reward encoder in the GRPO training. 
In contrast, the supervised refinement stage is applied only to the embedding model used for downstream matching, and does not affect the GRPO optimization process. 
Specifically, the refined embedding is used at inference time to compute similarity between encoded CoT and candidate predicates.

\textbf{Refinement Strategy.}
We perform supervised fine-tuning on the self-supervised embedding model using CoT-derived supervision. The training data is also restricted to base-set samples.
Specifically, the full CoT texts are used as queries, while the corresponding ground-truth predicates (possibly multiple per sample) serve as labels. The construction of the training data follows the same procedure as in Sec.~\ref{sec:c1_emb_unsupervised}, where image-level soft-balanced sampling is applied, and the dataset size is kept consistent.

\textbf{Multi-positive Contrastive Learning.}
We adopt a multi-positive InfoNCE objective for training. 
Given a mini-batch, we first aggregate all predicate labels across samples and remove duplicates to form a candidate predicate set. 
Each query is associated with multiple positive predicates within this set, while the remaining predicates are treated as negatives.
To improve computational efficiency, each unique predicate is encoded only once per batch. 
To prevent samples with a larger number of positives from dominating the training objective, we normalize the contributions of positive pairs by the number of positives for each query.

\textbf{Model and Training Details.}
The embedding encoder is initialized from the self-supervised model and further fine-tuned using LoRA, where the backbone parameters remain frozen while the LoRA adapters are initialized from the self-supervised stage and remain trainable. The LoRA configuration is kept consistent with the previous stage (rank $r=8$, scaling factor $\alpha=16$, and dropout $=0.1$).
Text representations are obtained using mean pooling. 
Training is performed with AdamW using a learning rate of $1\times10^{-4}$ and a cosine learning rate schedule with warmup.
Unless otherwise specified, we use a per-device batch size of 16 with gradient accumulation of 2, and train for 3 epochs. 
The temperature parameter is set to $\tau = 0.12$. We use the final checkpoint for the GRPO-optimized MLLM and select embedding checkpoints with the lowest loss on a validation subset held out from the training set.

\paragraph{SGDet Inference and Evaluation.}
We use the candidate object pairs selected by the upstream detector. Under graph-constraint evaluation, we retain only the highest-scoring predicate for each directed object pair and rank triplets solely by relation scores, without incorporating object detection confidence.

\paragraph{Predicate Templates.}
We use Gemini 3 to generate approximately 3--7 textual templates for each predicate, which are used to construct the multi-template predicate prototypes.

\subsection{Score Normalization and Prior Calibration}

During inference, we first compute the predicate retrieval score $s_k$ following Sec.~3.5. 
To reduce scale variation across different subject-object pairs, we apply pair-level z-score normalization over all predicate scores:
\begin{equation}
    \hat{s}_k = \frac{s_k-\mu}{\max(\sigma,\epsilon)},
\end{equation}
where $\mu$ and $\sigma$ are computed over all predicates for the current subject-object pair.

We then apply a lightweight subject-object prior calibration. For base predicates, the prior is estimated from training-set co-occurrence statistics with Dirichlet smoothing. For novel predicates, since no training annotations are available, we transfer the base prior according to predicate semantic similarity, where each novel predicate uses the mean prior of its top-$K$ nearest base predicates in the predicate embedding space. We set $K=5$ in all experiments.

The final calibrated score is:
\begin{equation}
    \tilde{s}_k =
    \hat{s}_k + \eta \log \big(q(p_k \mid c_i,c_j)),
\end{equation}
where $q$ denotes the base prior for base predicates and the transferred prior for novel predicates. We set $\eta=0.05$ in all experiments, and predict the relation by $\arg\max_k \tilde{s}_k$. This calibration is used only during inference.

\section{Supplementary Experiment: Zero-Shot PredCls Performance of Different MLLM Backbones}
\label{appendix:mllm_backbone}

\subsection{Experimental Setup}
This experiment is conducted to verify the impact of the basic vision-semantic alignment capability of the MLLM backbone on the zero-shot performance of open-vocabulary relation reasoning.

The four off-the-shelf backbones are evaluated without fine-tuning, using the same prompt, decoding strategy, and TP+PR retrieval pipeline. PR uses base-training statistics. Trained RaRe is included as a reference.

\subsection{Experimental Results}
We adopt the standard OV-SGG evaluation metrics, and report the global recall R@50/100 and category-averaged mean recall mR@50/100 on the full set (Total), base classes (Base), and novel classes (Novel). The results are summarized in Table \ref{tab:mllm_backbone}.

\begin{table}[h]
\centering
\caption{PredCls results of four off-the-shelf MLLM backbones using TP+PR without parameter updates, with trained RaRe included as a reference. PR uses base-training statistics.}
\resizebox{\textwidth}{!}{
\begin{tabular}{lcccccc}
\toprule
\textbf{MLLM Backbone} & \multicolumn{2}{c}{\textbf{Total}} & \multicolumn{2}{c}{\textbf{Base}} & \multicolumn{2}{c}{\textbf{Novel}} \\
& R@50/100 & mR@50/100 & R@50/100 & mR@50/100 & R@50/100 & mR@50/100 \\
\midrule
Qwen2.5-VL-3B-Instruct & 48.86/52.11 & 17.35/19.96 & 49.29/52.57 & 19.68/22.35 & 9.64/11.29 & 11.92/14.39 \\
Qwen3-VL-4B-Instruct & 52.90/55.68 & 20.68/23.54 & 53.40/56.21 & 23.37/25.93 & 12.71/13.67 & 14.39/17.95 \\
Qwen3-VL-8B-Instruct & 53.13/55.96 & 21.60/23.94 & 53.57/56.42 & 23.63/25.89 & 15.56/17.34 & 16.87/19.41 \\
Qwen3.5-4B & 53.22/56.08 & 21.72/23.96 & 53.69/56.58 & 23.75/26.30 & 13.22/14.34 & 16.98/18.50 \\
RaRe & 49.90/51.33 & 24.62/25.80 & 50.36/51.81 & 25.53/26.45 & 12.75/13.60 & 22.49/24.27 \\
\bottomrule
\end{tabular}
}
\label{tab:mllm_backbone}

\end{table}

The experimental results show that the MLLM backbone with stronger basic vision-semantic understanding capability exhibits better comprehensive performance on the zero-shot open-vocabulary relation reasoning task, especially stronger generalization ability on unseen novel categories. This result provides a basic reference for the backbone selection and optimization of subsequent MLLM-based OV-SGG methods.

\section{Sensitivity analysis}
\label{appendix:Sensitivity analysis}

All experiments in this section are conducted on the same trained RaRe checkpoint under the PredCls setting. We use a fixed configuration of $K=5$ and $\eta=0.05$, chosen before test-set evaluation.

\subsection{Sensitivity to the Number of Transferred Predicates}

The sensitivity to the number of transferred base predicates is investigated by varying $K \in \{1,3,5,10\}$, with the results summarized in Table \ref{tab:K sensitive}. The optimal Total and Novel mR are attained at $K=5$, whereas performance remains largely stable for $K \in \{3,5,10\}$. The degradation observed at $K=1$ suggests that the prior transferred from a single neighbor is more susceptible to semantic mismatch, and is therefore less reliable for calibration.

\begin{table}[ht]
\centering
\small
\caption{Sensitivity analysis of the number of transferred base predicates $K$. All results are reported under the PredCls setting with $\eta=0.05$ and uniform averaging transfer.}
\begin{tabular}{lccc}
\toprule
$K$ & Total mR@50/100 & Base mR@50/100 & Novel mR@50/100 \\
\midrule
1  & 22.86/24.53 & 24.94/26.60 & 18.02/19.72 \\
3  & 23.61/25.31 & 24.80/26.48 & 20.82/22.58 \\
5  & \textbf{24.62/25.80} & \textbf{25.53}/26.45 & \textbf{22.49/24.27} \\
10 & 23.95/25.48 & 24.94/\textbf{26.69} & 21.65/22.65 \\
\bottomrule
\end{tabular}
\label{tab:K sensitive}

\end{table}

\subsection{Sensitivity to the Prior Weight}

The sensitivity to the subject--object prior weight $\eta$ in Eq.~(12) is further examined, with the results reported in Table \ref{tab:weight sensitive}. The results indicate that the prior should be retained as a low-weight calibration signal: a modest weight yields a more favorable overall balance, whereas an excessive weight ($\eta=0.1$) leads to a substantial degradation in Novel mR. This observation substantiates the treatment of the prior as a lightweight calibration term, such that relation-semantic scores remain the dominant prediction signal. Among the tested values, $\eta=0.01$ achieves the highest Novel mR, whereas the predefined setting $\eta=0.05$ achieves the highest Total and Base mR.

\begin{table}[ht]
\centering
\small
\caption{Sensitivity analysis of the subject--object prior weight $\eta$. All results are reported under the PredCls setting with $K=5$ and uniform averaging transfer.}
\begin{tabular}{lccc}
\toprule
$\eta$ & Total mR@50/100 & Base mR@50/100 & Novel mR@50/100 \\
\midrule
0    & 23.03/24.33 & 22.12/23.32 & 25.15/26.71 \\
0.01 & 24.50/25.65 & 23.54/24.85 & \textbf{26.75/27.50} \\
0.05 & \textbf{24.62/25.80} & \textbf{25.53/26.45} & 22.49/24.27 \\
0.1  & 19.39/21.06 & 23.03/24.79 & 10.92/12.37 \\
\bottomrule
\end{tabular}
\label{tab:weight sensitive}

\end{table}

\subsection{Comparison of Prior Transfer Strategies}

Table \ref{tab:transfer} compares uniform averaging with distance-weighted transfer and with the removal of novel-prior transfer. Without transfer, base priors are retained and novel priors are fixed to $0.05/15$. This penalizes novel scores, reducing Novel mR@50 to 2.33 and highlighting the sensitivity to base--novel calibration.

\begin{table}[ht]
\centering
\small
\caption{Comparison of novel-predicate prior transfer strategies. All results are reported under the PredCls setting with $K=5$ and $\eta=0.05$.}
\begin{tabular}{lccc}
\toprule
Method & Total mR@50/100 & Base mR@50/100 & Novel mR@50/100 \\
\midrule
Uniform & \textbf{24.62/25.80} & 25.53/26.45 & \textbf{22.49/24.27} \\
Distance-weighted & 23.60/25.30 & 24.82/26.56 & 20.73/22.35 \\
No novel-prior transfer & 18.99/20.13 & \textbf{26.14/27.56} & 2.33/2.78 \\
\bottomrule
\end{tabular}
\label{tab:transfer}
\end{table}

\subsection{Examples of Transferred Neighbors}

Table \ref{tab:transfer neighbour} presents examples of novel-prior transfer under the VG150 PredCls split, with neighbors retrieved exclusively from its base predicate vocabulary

\begin{table}[ht]
\centering
\small
\caption{Examples of the top-5 nearest base predicates used for novel-prior transfer under the VG150 PredCls split.}
\begin{tabular}{@{}ll@{}}
\toprule
Novel predicate & Top-5 nearest base predicates \\
\midrule
across      & 1) to, 2) over, 3) in front of, 4) against, 5) behind \\
laying on   & 1) on, 2) sitting on, 3) under, 4) covered in, 5) attached to \\
on back of  & 1) on, 2) behind, 3) against, 4) over, 5) attached to \\
\bottomrule
\end{tabular}
\label{tab:transfer neighbour}
\end{table}

\section{Comparison with Alternative MLLM Optimization Strategies}
\label{appendix:sftvsgrpo}

To compare optimization strategies under an identical backbone, an SFT-only checkpoint and a Relation-R1-style GRPO checkpoint were trained with same Qwen3-VL-4B-Instruct and evaluated under the same protocol. The SFT baseline is supervised with ground-truth base predicates. The GRPO baseline adopts a simplified form of the Relation-R1 reward~\cite{li2025relationr1progressivelycognitivechainofthought}: since the PredCls setting provides object labels and boxes and queries a single object pair per sample, the original triplet-level criterion reduces to predicate correctness,
$$
r = r_{\mathrm{format}} + r_{\mathrm{pred}},
$$
where $r_{\mathrm{pred}} = 1$ if the predicted predicate matches the ground-truth predicate and
$0$ otherwise.

Since only base-predicate annotations are available under the standard OV-SGG protocol, both SFT and correctness-based GRPO directly optimize a finite base-label space. This improves base-category recognition but biases the model toward base predicates and weakens novel-predicate generalization (Table \ref{tab:sftgrpo}).

RaRe instead avoids predicate correctness during policy optimization and optimizes object-pair specificity in the relational representation space, which prevents direct attraction toward the base-label space. The stronger and more balanced mR of RaRe indicates additional benefits over conventional SFT and correctness-based GRPO.

\begin{table}[ht]
\centering
\footnotesize
\setlength{\tabcolsep}{3pt}
\caption{Same-backbone comparison of MLLM optimization strategies under the PredCls setting on
the full VG150 test set. GRPO uses a binary predicate-correctness reward adapted from Relation-R1, while RaRe uses the specificity reward in Eq.~(8).}
\begin{tabular}{lcccccc}
\toprule
& \multicolumn{2}{c}{Total (Relation)} & \multicolumn{2}{c}{Base (Relation)} & \multicolumn{2}{c}{Novel (Relation)} \\
\cmidrule(lr){2-3}\cmidrule(lr){4-5}\cmidrule(lr){6-7}
Model & R@50/100 & mR@50/100 & R@50/100 & mR@50/100 & R@50/100 & mR@50/100 \\
\midrule
SFT & \textbf{55.92/58.49} & 16.83/18.70 & \textbf{56.53/59.13} & 20.69/22.80 & 7.33/8.13 & 7.84/9.14 \\
GRPO & 55.05/57.64 & 20.18/22.52 & 55.60/58.22 & 23.56/26.03 & 10.34/10.97 & 12.30/14.33 \\
RaRe (Ours) & 49.90/51.33 & \textbf{24.62/25.80} & 50.36/51.81 & \textbf{25.53/26.45} & \textbf{12.75/13.60} & \textbf{22.49/24.27} \\
\bottomrule
\end{tabular}
\label{tab:sftgrpo}
\end{table}

\section{Supplementary Experiment: SGCls Setting Results}
\label{appendix:sgcls}
In this section, we provide the complete experimental results of our method under the SGCls (Scene Graph Classification) setting, as supplementary validation of our method's robustness.

\begin{table*}[ht]
\centering
\caption{\textbf{Experimental results of OVR-SGG on SGCls setting.}}
\resizebox{\linewidth}{!}{
\begin{tabular}{@{}l l c c c c c@{}}
\toprule
{\textbf{Model}} & \multicolumn{2}{c}{\textbf{Total (Relation)}} & \multicolumn{2}{c}{\textbf{Base (Relation)}} & \multicolumn{2}{c}{\textbf{Novel (Relation)}} \\
\cmidrule(lr){2-3} \cmidrule(lr){4-5} \cmidrule(l){6-7}
&  R@50/100 & mR@50/100 & R@50/100 & mR@50/100 & R@50/100 & mR@50/100 \\
\midrule
SGTR+ & 21.42/26.60 & 9.13/11.58 & 32.18/36.11 & 13.12/16.88 & 0.00/0.00 & 0.00/0.00 \\
SGTR+ + RAHP & 26.65/30.83 & 12.98/\textbf{16.81} & 32.63/36.83 & 13.86/\textbf{17.14} & 14.37/\textbf{21.72} & 9.91/15.04 \\
ACC+Ours & \textbf{30.46}/\textbf{33.56} & \textbf{13.76}/16.62 & \textbf{33.79}/\textbf{37.39} & \textbf{14.30}/17.02 & \textbf{15.50}/17.38 & \textbf{12.51}/\textbf{15.67} \\
%$^\dagger$
\bottomrule
\end{tabular}
}
\label{tab:sgcls_results}

\end{table*}
Under the SGCls setting, our method achieves 12.51\% on Novel mR@50, outperforming the strongest baseline SGTR+ + RAHP \cite{liu2025relationawarehierarchicalpromptopenvocabulary} (9.91\%).

\section{Cross-Task Transfer: Training-Free Zero-Shot HOI}
\label{appendix:zs_hoi}
We examine whether our relation reasoning pipeline can be adapted to zero-shot HOI using an off-the-shelf MLLM without task-specific training or parameter updates.

\subsection{Experimental Setup}
We use the off-the-shelf Qwen3-VL-4B-Instruct and Qwen3-VL-Embedding-2B models without SGG- or HOI-specific fine-tuning.
Human--object pair construction and object-conditioned HOI candidate filtering follow the training-free configuration of ADA-CM~\cite{lei2023efficient}, using its DETR-R50 detector.
We apply our CoT-based embedding-retrieval pipeline to score the candidate interactions.

Experiments are conducted on HICO-DET \cite{chao2018learning}, the standard benchmark for ZS-HOI. We follow the general ZS-HOI protocol and use mean Average Precision (mAP) as the evaluation metric, reporting results on the full set (Full), rare categories (Rare), and non-rare categories (Non-rare). We compare our method with state-of-the-art ZS-HOI methods under the exact same training-free setting, including UPT (TF)~\cite{zhang2022efficient}, ADA-CM \cite{lei2023efficient} (ICCV 2023), and DA-HOI \cite{xuan2026zero} (ICLR 2026).

\subsection{Results and Conclusion}
The experimental results are summarized in Table \ref{tab:zs_hoi_results}.

\begin{table}[ht]
\centering
\caption{Training-Free Zero-Shot HOI results on HICO-DET (metric: mAP). All methods are under the fully training-free setting.}
\begin{tabular}{lccc}
\toprule
\textbf{Method} & \textbf{Full} & \textbf{Rare} & \textbf{Non-rare} \\
\midrule
UPT (TF) & 13.64 & 9.66 & 14.85 \\
ADA-CM (ICCV 2023) & 25.19 & 27.24 & 24.58 \\
DA-HOI (ICLR 2026) & \textbf{31.50} & - & - \\
\textbf{Ours} & 27.71 & \textbf{29.06} & \textbf{27.31} \\
\bottomrule
\end{tabular}
\label{tab:zs_hoi_results}

\end{table}

Our pipeline achieves 27.71 Full mAP and 29.06 Rare mAP on HICO-DET without task-specific training.These results provide evidence of cross-task transfer to zero-shot HOI.

\section{Additional Analysis on Representation Properties}

\label{appendix:Additional Analysis on Representation Properties}

\begin{figure}[h]
    \centering
        \centering
        \includegraphics[width=\linewidth]{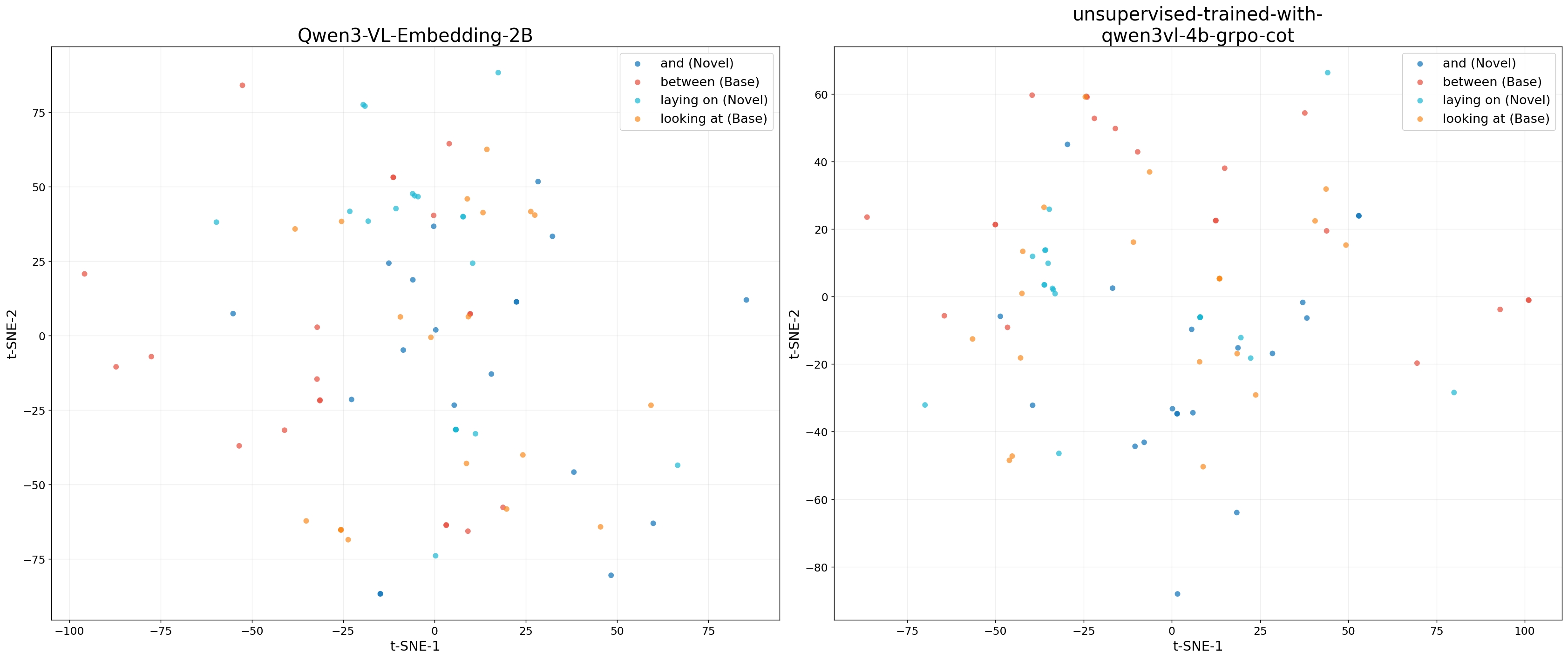}
        \small
        \caption{t-SNE visualization of CoT representations projected by Baseline (Qwen3-VL-Embedding-2B), and +CoT+GRPO.}
        \label{fig:cot_embedding}
\end{figure}

\begin{table}[ht]
\centering
\caption{\textbf{Discriminability of Unsupervised-trained Embedding} We report Silhouette Score and macro\_mean\_ratio for Global, Base, and Novel subsets. $\uparrow\Delta$ denotes improvement over the baseline. Baseline uses the off-the-shelf embedding model and CoTs from the unoptimized MLLM. +CoT replaces only the embedding model with its SSL-trained counterpart. +CoT+GRPO retains the same SSL-trained embedding model and uses CoTs from the GRPO-optimized MLLM.}
\label{tab:embedding_analysis}
\small
\centering
\begin{tabular}{@{} l c c c @{}}
\toprule
\textbf{Metric} & \textbf{Baseline} & \textbf{+CoT} & \textbf{+CoT+GRPO} \\
\midrule
Silhouette (Global) & -0.1313 & -0.0935 ($\uparrow$0.0378) & -0.0924 ($\uparrow$0.0389) \\
Silhouette (Base)   & -0.1040 & -0.0689 ($\uparrow$0.0351) & -0.0657 ($\uparrow$0.0383) \\
Silhouette (Novel)  & +0.0298 & +0.0514 ($\uparrow$0.0216) & +0.0511 ($\uparrow$0.0213) \\
\midrule
macro\_mean\_ratio (Global) & 0.0944 & 0.2876 ($\uparrow$0.1932) & 0.3177 ($\uparrow$0.2233) \\
macro\_mean\_ratio (Base)   & 0.0863 & 0.2766 ($\uparrow$0.1903) & 0.3071 ($\uparrow$0.2208) \\
macro\_mean\_ratio (Novel)  & 0.9683 & 1.4294 ($\uparrow$0.4611) & 1.4717 ($\uparrow$0.5034) \\
\bottomrule
\end{tabular}
\end{table}

\paragraph{Discriminability Analysis}
Since qualitative clustering visualizations in Figure\ref{fig:cot_embedding} are insufficient to capture fine-grained inter-predicate variation, we adopt two quantitative metrics: Silhouette Score, which measures intra-class compactness relative to inter-class separation (range $[-1,1]$, higher is better), and \textbf{macro \_mean \_ratio}, the weighted mean of nearest inter-class distance over intra-class scatter per category (values $>1$ indicate basic discriminability). Both metrics are calculated on a dedicated evaluation subset of 2,000 images containing 12,686 relational triplets, spanning both base and novel predicates. We utilize CoT representations generated by Qwen3-VL-4B-Instruct across different embedding models to assess how our training strategy enhances the relational separability in the embedding space, as summarized in Table~\ref{tab:embedding_analysis}.

Both trained variants improve over the baseline on all reported metrics.
Compared with CoT-only training, the GRPO variant further improves most metrics, while Novel Silhouette remains nearly unchanged (0.0514 versus 0.0511).
Global and Base Silhouette scores remain negative, indicating improved but still limited cluster separation.
Overall, these results suggest improved relational discriminability, including for novel predicates.

The metrics are defined as follows:

(1) \textit{Silhouette Score} for a sample $i$ is given by:
\begin{equation}
s(i) = \frac{b(i) - a(i)}{\max\{a(i), b(i)\}}
\end{equation}
where $a(i) = \frac{1}{|C_I| - 1} \sum_{j \in C_I, i \neq j} d(i, j)$ is the average intra-class distance, and $b(i) = \min_{J \neq I} ( \frac{1}{|C_J|} \sum_{j \in C_J} d(i, j) )$ is the average distance to the nearest neighboring cluster.

(2) \textbf{macro\_mean\_ratio} for a set of predicates is defined as:
\begin{equation}
\text{macro\_mean\_ratio} = \frac{\sum_{k=1}^{K} n_k \cdot R_k}{\sum_{k=1}^{K} n_k}, \quad R_k = \frac{d_{cos}(\mu_k, \mu_{nearest})}{\sigma_k + \tau}
\end{equation}
where $n_k$ is the number of samples for predicate $C_k$, and $R_k$ is the stabilized discriminability ratio. $d_{cos}(\mu_k, \mu_{nearest})$ denotes the cosine distance between the centroid of $C_k$ and its nearest neighbor centroid, while $\sigma_k$ is the mean cosine distance of samples in $C_k$ to their own centroid. To prevent divergence when $\sigma_k \to 0$, we introduce a stabilization term $\tau = \max(10^{-6}, 0.1 \times \text{median}(\{\sigma\}_K))$. Despite its name, macro\_mean\_ratio is weighted by category frequency rather than averaged equally across categories.

\paragraph{CoT Length}
Table~\ref{tab:cot_length} compares the average CoT length on the full VG150 test set under PredCls.
RaRe reduces the average length from 392.8 to 297.1 tokens, suggesting that its gains do not rely on longer generated rationales.

\begin{table}[ht]
\centering
\caption{\textbf{Average CoT length on the full VG150 test set under PredCls.}}
\label{tab:cot_length}
\small
\begin{tabular}{@{}lc@{}}
\toprule
\textbf{Model} & \textbf{Average Length (tokens)} \\
\midrule
Qwen3-VL-4B-Instruct & 392.8 \\
RaRe & 297.1 \\
\bottomrule
\end{tabular}
\end{table}

\section{Additional Controls and Inference Cost}
\label{appendix:additional exp}

\textbf{Object-only control.} To verify that the performance relies on visual input rather than on
object annotations alone, the checkpoint is re-evaluated with the image removed, while object
labels, boxes, prompt, and inference pipeline are kept unchanged (Table \ref{tab:wo image}). Performance drops
consistently across all three groups, and the drop is largest on novel predicates (from 22.49 to
13.71 at mR@50), indicating that object annotations alone do not account for the reported results.

\textbf{Direct output vs. embedding retrieval.} Using the same off-the-shelf Qwen3-VL-4B-Instruct
model and prompt, direct predicate prediction is compared with the training-free
embedding-retrieval pipeline (Table \ref{tab:direct output}). Retrieval improves Total, Base, and Novel mR@100 by 8.05,
5.77, and 13.37 points, respectively. Since the MLLM and the prompt are identical in both
variants, the gap is attributable to the embedding-based retrieval, which also serves as the
reward encoder in Stage~2 rather than being introduced only at inference time.

\paragraph{Fixed-proposal SGDet control.}
Table~\ref{tab:vg_fixed_proposal} compares two configurations using identical ACC boxes, object labels, candidate pairs, decoding settings, candidate filtering, and TP-based retrieval.
Adding RL, SSL, and PR yields modest Total mR@50/100 gains of 0.30/0.13 points, while reducing Novel mR@50/100 by 3.52/4.25 points. Together with the PSG ablation in Table~\ref{tab:psg_ablation}, this indicates that the benefits of learning and calibration vary across evaluation settings.
\begin{table}[t]
\centering
\caption{\textbf{Fixed-proposal SGDet control on the full VG150 test set.}
Both configurations use identical ACC object proposals and candidate pairs.
The RaRe configuration adds RL, SSL, and PR to training-free retrieval; ST is not used.}
\label{tab:vg_fixed_proposal}
\small
\begin{tabular}{@{}lcc@{}}
\toprule
\textbf{Configuration}
& \textbf{Total mR@50/100}
& \textbf{Novel mR@50/100} \\
\midrule
Training-free retrieval & 10.08/12.19 & 10.21/12.27 \\
+ RL + SSL + PR & 10.38/12.32 & 6.69/8.02 \\
\bottomrule
\end{tabular}
\end{table}

\textbf{Inference cost.} Table \ref{inference time} reports the per-image latency of the SGDet pipeline on four
NVIDIA L40 GPUs. CoT generation dominates the cost (3.526s, 64.7\%), followed by object detection
(1.247s, 22.9\%), while embedding retrieval accounts for 0.677s.

\begin{table}[ht]
\centering
\small
\caption{Object-only control under the PredCls setting on the full VG150 test set. Only the image
is removed; object labels, boxes, prompt, and inference pipeline are unchanged.}
\begin{tabular}{lccc}
\toprule
Variant & Total mR@50/100 & Base mR@50/100 & Novel mR@50/100 \\
\midrule
RaRe w/o image & 16.78/18.02 & 18.10/19.50 & 13.71/14.56 \\
RaRe & \textbf{24.62/25.80} & \textbf{25.53/26.45} & \textbf{22.49/24.27} \\
\bottomrule
\end{tabular}
\label{tab:wo image}
\end{table}

\begin{table}[ht]
\centering
\small
\caption{Training-free comparison between direct MLLM output and embedding retrieval under the
PredCls setting on the full VG150 test set. Both variants use the same MLLM and prompt.}
\begin{tabular}{lccc}
\toprule
Method & Total mR@100 & Base mR@100 & Novel mR@100 \\
\midrule
Direct MLLM output & 12.20 & 13.25 & 9.76 \\
Embedding retrieval & 20.25 & 19.02 & 23.13 \\
\midrule
Improvement & +8.05 & +5.77 & +13.37 \\
\bottomrule
\end{tabular}
\label{tab:direct output}
\end{table}

\begin{table}[ht]
\centering
\small
\caption{Inference-time breakdown of the SGDet pipeline on four NVIDIA L40 GPUs, reported per
image.}
\begin{tabular}{lrr}
\toprule
Component & Time (s/image) & Share \\
\midrule
Object detection (ACC) & 1.247 & 22.9\% \\
CoT reasoning generation & 3.526 & 64.7\% \\
Embedding retrieval & 0.677 & 12.4\% \\
\midrule
Total & 5.450 & 100.0\% \\
\bottomrule
\end{tabular}
\label{inference time}
\end{table}

\section{Long-Tailed Recognition and Novel-Predicate Coverage under SGDet}
\label{appendix:acccompare}

\paragraph{Base-Predicate Long-Tailed Recognition.}
To distinguish training-frequency effects from novel-predicate generalization, we rank the 35 base predicates by their frequency in the VG150 training set and divide them into head, body, and tail groups using a 20\%/40\%/40\% class split.
Both methods are evaluated on the full VG150 test set with identical ACC proposals under graph-constraint evaluation.
Table~\ref{tab:base_long_tail} shows that RaRe substantially improves body and tail recognition, with a small decrease on head predicates.
Under the same top-100 prediction budget, it also allocates larger shares of base predictions to body and tail predicates (Table~\ref{tab:base_prediction_distribution}).
The accompanying recall gains indicate that this broader coverage translates into improved recognition of less frequent predicates.

\begin{table}[ht]
\centering
\caption{\textbf{Frequency-stratified results on VG150 base predicates under SGDet.}
Head, body, and tail contain 7, 14, and 14 predicates, respectively, ranked by training frequency.}
\label{tab:base_long_tail}
\small
\begin{tabular}{@{}lccc@{}}
\toprule
Method & Head mR@50/100 & Body mR@50/100 & Tail mR@50/100 \\
\midrule
ACC  & \textbf{24.0/28.0} & 0.7/0.9 & 0.0/0.0 \\
RaRe & 22.8/26.7 & \textbf{14.3/16.3} & \textbf{4.2/5.7} \\
\bottomrule
\end{tabular}
\end{table}

\begin{table}[ht]
\centering
\caption{\textbf{Prediction distribution across base-predicate frequency groups.}
Base pred.\ denotes the share of all top-100 predictions assigned to base predicates.
Head, body, and tail shares are normalized within base predictions.}
\label{tab:base_prediction_distribution}
\small
\setlength{\tabcolsep}{5pt}
\begin{tabular}{@{}lcccc@{}}
\toprule
Method & Base pred.\ (\%) & Head (\%) & Body (\%) & Tail (\%) \\
\midrule
ACC  & 64.36 & 99.4 & 0.6 & $\approx 0.0$ \\
RaRe & 70.07 & 72.9 & 19.2 & 7.9 \\
\bottomrule
\end{tabular}
\end{table}
\paragraph{Novel-Predicate Coverage.}

All statistics are computed on the full VG test set with the fixed ACC proposals of the SGDet evaluation in Table~\ref{tab:sgdet_results}, where Ours denotes the RaRe variant. The 15 OvR novel predicates are ranked by ground-truth frequency and split 20\%/40\%/40\% into three head predicates (on, of, and in), six body predicates, and six tail predicates; the head group holds 92.4\% of the novel GT instances, the body group 6.9\%, and the tail group 0.7\%. Since mR averages over predicates, each of the 12 body and tail predicates weighs as much as on.\\

Table~\ref{tab:compare acc1}  reports the prediction counts per group: 942k, 0, and 124 for ACC and 644k, 77k, and 70k for RaRe, with an identical total of 2,644,600 predictions and a smaller novel share for RaRe (29.93\% vs. 35.64\%). At mR@50, ACC obtains 0.124 on the head group and 0.000 on the other two, whereas RaRe obtains 0.083, 0.062, and 0.064. Per predicate (Table~\ref{tab:caompare acc2}), RaRe is lower on on (0.103 vs. 0.273) and higher on of (0.078 vs. 0.072), in (0.068 vs. 0.027), and 7 of the 12 body and tail predicates, including riding (0.310), eating (0.134), and says (0.250). \\

Five predicates stay at zero for both models at R@50 (wears, part of, walking in, painted on, and playing; part of reaches 0.004 for RaRe at R@100), and some of them receive far more predictions than they have GT instances (says: 12 GT instances, 44k predictions), so their recall rests on few positives. The gain thus comes from categories that ACC leaves at zero, each worth 1/15 of the mR, whereas the loss on on is bounded by a single 1/15; the remaining gap is set by the five predicates that neither model recovers.

\begin{table}[ht]
\centering
\footnotesize
\setlength{\tabcolsep}{3pt}
\caption{Novel-predicate recall and prediction counts by head/body/tail group under SGDet with
fixed ACC proposals. All results are reported at mR@50/100 on the full VG test set.}
\begin{tabular}{lcccccc}
\toprule
Bucket & GT share & Ours@50 & ACC@50 & Ours@100 & ACC@100 & Pred.\ (Ours/ACC) \\
\midrule
Head & 92.4\% & 0.083 & 0.124 & 0.112 & 0.159 & 644k/942k \\
Body & 6.9\% & 0.062 & 0.000 & 0.074 & 0.000 & 77k/0 \\
Tail & 0.7\% & 0.064 & 0.000 & 0.070 & 0.000 & 70k/124 \\
\bottomrule
\end{tabular}
\label{tab:compare acc1}
\end{table}

\begin{table}[ht]
\centering
\footnotesize
\setlength{\tabcolsep}{3pt}
\caption{Per-predicate novel recall, test-set frequency, and prediction counts under the VG150 SGDet split with fixed ACC proposals. Pred.\ \# reports the number of predictions, and paired entries correspond to Ours/ACC}
\begin{tabular}{llr cccc rr}
\toprule
& & & \multicolumn{2}{c}{R@50} & \multicolumn{2}{c}{R@100} & \multicolumn{2}{c}{Prediction (Ours/ACC)} \\
\cmidrule(lr){4-5}\cmidrule(lr){6-7}\cmidrule(lr){8-9}
Predicate & Bucket & Test\# & Ours & ACC & Ours & ACC & \# & \% \\
\midrule
on & Head & 50947 & 0.103 & 0.273 & 0.119 & 0.334 & 409k/827k & 15.47/31.26 \\
of & Head & 17074 & 0.078 & 0.072 & 0.129 & 0.110 & 81k/95k & 3.06/3.58 \\
in & Head & 9138 & 0.068 & 0.027 & 0.089 & 0.031 & 154k/21k & 5.82/0.79 \\
\midrule
wears & Body & 2059 & 0.000 & 0.000 & 0.000 & 0.000 & 1.7k/0 & 0.06/0.00 \\
riding & Body & 1348 & 0.310 & 0.000 & 0.352 & 0.000 & 26k/0 & 0.99/0.00 \\
attached to & Body & 742 & 0.036 & 0.000 & 0.058 & 0.000 & 21k/0 & 0.78/0.00 \\
belonging to & Body & 666 & 0.007 & 0.000 & 0.015 & 0.000 & 8.6k/0 & 0.33/0.00 \\
walking on & Body & 577 & 0.006 & 0.000 & 0.009 & 0.000 & 11k/0 & 0.40/0.00 \\
for & Body & 335 & 0.010 & 0.000 & 0.010 & 0.000 & 9.2k/0 & 0.35/0.00 \\
\midrule
eating & Tail & 232 & 0.134 & 0.000 & 0.168 & 0.000 & 6.6k/0 & 0.25/0.00 \\
part of & Tail & 138 & 0.000 & 0.000 & 0.004 & 0.000 & 8.1k/0 & 0.31/0.00 \\
walking in & Tail & 105 & 0.000 & 0.000 & 0.000 & 0.000 & 5.2k/1 & 0.20/0.00 \\
painted on & Tail & 96 & 0.000 & 0.000 & 0.000 & 0.000 & 2.2k/0 & 0.08/0.00 \\
playing & Tail & 28 & 0.000 & 0.000 & 0.000 & 0.000 & 3.8k/0 & 0.15/0.00 \\
says & Tail & 12 & 0.250 & 0.000 & 0.250 & 0.000 & 44k/123 & 1.68/0.00 \\
\bottomrule
\end{tabular}
\label{tab:caompare acc2}
\end{table}

\clearpage
\section{More Qualitative Comparison Result}
\label{appendix:compare}
Predictions are obtained by embedding retrieval for the queried object pair and may differ from the generated summary. Correctness is determined by matching the ground-truth predicate; a correct prediction does not imply that every statement in the rationale is visually accurate.

\begin{figure}[htbp]
    \centering
    
    % Sample 1
    \qualitativeSample{
        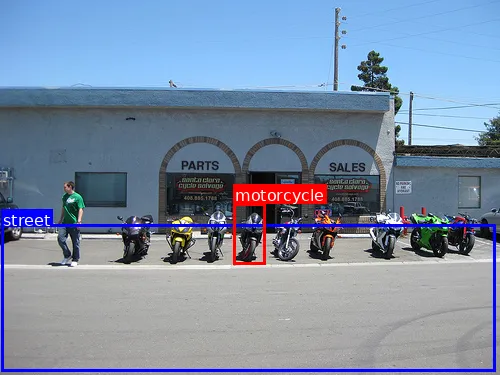 % 替换为你的图片路径
    }{$<$VISUAL\_EVIDENCE$>$ 
The image shows a row of motorcycles parked on a paved street in front of a building with "PARTS" and "SALES" signs. The motorcycles are aligned parallel to the street, positioned on the asphalt surface. The street occupies the lower portion of the image, extending from the left edge to the right, with the motorcycles resting directly on it. The building is behind the motorcycles, and the street is clearly the ground surface upon which the motorcycles are situated. 
$</$VISUAL\_EVIDENCE$>$

$<$REASONING\_THINKING$>$ 
Step 1: Identify the objects. The two objects are a motorcycle and a street. The motorcycle is visible in the foreground, parked on the street. 
Step 2: Analyze spatial relationship. The motorcycle is positioned on the surface of the street, which is a paved road. The street is the ground plane, and the motorcycle is resting on it. 
Step 3: Determine interaction mechanism. The motorcycle is not floating or suspended; it is placed on the street surface, indicating a physical resting relationship. The street provides a stable, flat surface for the motorcycle to be parked on. 
Step 4: Confirm no other relationships. There is no indication of the motorcycle being attached to the building, or the street being above the motorcycle. The motorcycle is clearly on the street, not beside or behind it. 
Step 5: Conclude relationship. The motorcycle is resting on the street, which is the most direct and logical spatial relationship based on the visual evidence. 
$</$REASONING\_THINKING$>$

$<$FINAL\_SUMMARY$>$ 
motorcycle on street
The motorcycle is parked on the street surface.
$</$FINAL\_SUMMARY$>$
    }{
        motorcycle on street
    }{
        $<$VISUAL\_EVIDENCE$>$  
The image shows a motorcycle positioned on a paved surface, which appears to be a street. The motorcycle is stationary, with its wheels in contact with the ground, and it is located near the edge of the street. The street is a long, flat expanse extending horizontally across the frame, and the motorcycle is clearly placed on it rather than on a sidewalk or other surface. The surrounding area suggests an outdoor urban or suburban setting.
$</$VISUAL\_EVIDENCE$>$

$<$REASONING\_THINKING$>$  
The motorcycle is resting on the surface of the street, with no visible signs of motion or active use. Its wheels are aligned with the street’s edge, and there is no indication of it being in motion. The position and orientation of the motorcycle suggest it is not currently being driven. The lack of movement and the environment imply that the motorcycle is not operating at the moment. The street provides a flat, stable surface suitable for the motorcycle to be parked on. The interaction between the motorcycle and the street is consistent with a scenario where the motorcycle is temporarily placed in a stationary position.
$</$REASONING\_THINKING$>$

$<$FINAL\_SUMMARY$>$  
motorcycle parked on street  
The motorcycle is stationary on the surface of the street, indicating it is not in motion and is instead resting on the street.
$</$FINAL\_SUMMARY$>$
    }{
        motorcycle parked on street
    }

    \label{fig:qualitative_results}
\end{figure}

\begin{figure}[htbp]
    \centering
    
    % Sample 1
    \qualitativeSample{
        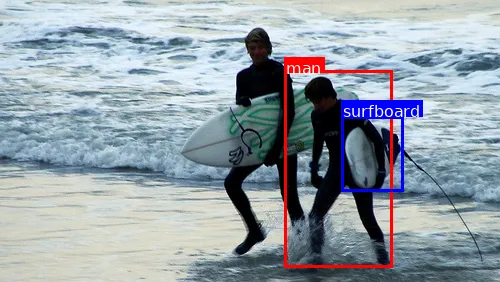 % 替换为你的图片路径
    }{$<$VISUAL\_EVIDENCE$>$
The image shows two individuals in wetsuits standing in shallow ocean water, with waves breaking behind them. The man on the left is holding a surfboard under his arm, which is predominantly white with green and black markings. The surfboard is positioned diagonally across his body, extending from his left shoulder to his right hip. The second individual, to the right, is also holding a surfboard, which is partially obscured by his body and appears to be tucked under his left arm. The surfboard he holds is mostly white with a visible black leash extending toward the right side of the frame. Both individuals are facing forward, with their bodies angled slightly toward the camera, and are partially submerged in water, creating splashes around their legs.
$</$VISUAL\_EVIDENCE$>$

$<$REASONING\_THINKING$>$
Step 1: Identify the two objects and their respective holders. The man on the left is holding a surfboard, and the second individual is also holding a surfboard. Both are in wetsuits, suggesting they are surfers.
Step 2: Analyze the spatial relationship. The surfboard held by the man on the left is clearly visible and positioned under his arm, with its length extending diagonally across his torso. The surfboard held by the second individual is similarly held under the arm, though partially obscured. Both surfboards are being carried, not actively used for surfing at this moment.
Step 3: Determine the interaction mechanism. The interaction is one of carrying or transporting the surfboards. The surfboards are not in use for riding waves but are being carried by the individuals, likely preparing to enter the water or having just exited.
Step 4: Consider context. The presence of ocean waves and the individuals’ wetsuits reinforce the context of surfing. The fact that both are holding surfboards and standing in shallow water suggests they are either preparing to surf or have just finished.
Step 5: Conclude the relationship. The predicate relationship is that the man is carrying a surfboard, and the second individual is also carrying a surfboard. The relationship is mutual and symmetric in nature, as both are engaged in the same action of carrying surfboards.
$</$REASONING\_THINKING$>$

$<$FINAL\_SUMMARY$>$
man carries surfboard, second individual carries surfboard. Both individuals are carrying surfboards while standing in shallow ocean water, likely preparing for or having just finished surfing.
$</$FINAL\_SUMMARY$>$

    }{
        man on surfboard
    }{
$<$VISUAL\_EVIDENCE$>$  
The image shows a man standing near a surfboard. The surfboard is positioned horizontally and appears to be held or carried by the man, with one end near his body and the other extending outward. The man is facing forward, and the surfboard is aligned parallel to his torso. The surfboard's orientation suggests it is being carried rather than being placed on the ground or used for surfing.
$</$VISUAL\_EVIDENCE$>$

$<$REASONING\_THINKING$>$  
The man is holding the surfboard in a way that indicates he is transporting it. The surfboard is not floating or being ridden, nor is it placed on the ground. Its position relative to the man—aligned with his body and held close—suggests it is being carried. This indicates an interaction where the man is moving the surfboard with him, likely for transport to or from a beach. The alignment and grip imply a structured method of carrying, consistent with a person holding an object.
$</$REASONING\_THINKING$>$

$<$FINAL\_SUMMARY$>$  
man carrying surfboard  
The man is transporting the surfboard by holding it horizontally along his body.
$</$FINAL\_SUMMARY$>$
    }{
        man carrying surfboard 
    }

    \label{fig:qualitative_results}
\end{figure}

\begin{figure}[htbp]
    \centering
    
    % Sample 1
    \qualitativeSample{
        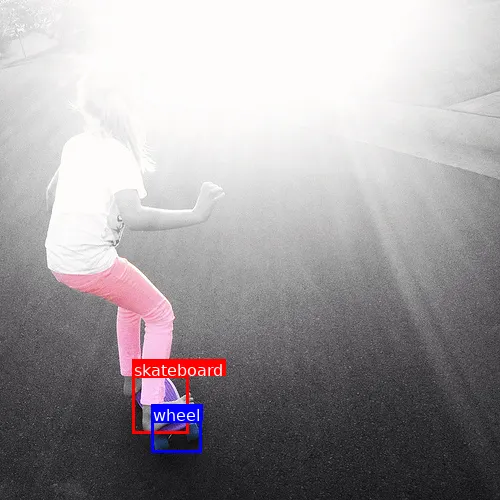 % 替换为你的图片路径
    }{$<$VISUAL\_EVIDENCE$>$
The skateboard is positioned beneath the person’s feet, with the person standing on it and actively riding. The wheel is one of the four wheels attached to the underside of the skateboard, visible in the lower portion of the skateboard’s structure. The skateboard is oriented horizontally on the asphalt surface, and the wheel is part of this assembly, directly connected to the board’s deck. The image is taken from behind the rider, with strong backlighting creating a bright flare, but the skateboard and wheel remain clearly identifiable in the foreground.
$</$VISUAL\_EVIDENCE$>$

$<$REASONING\_THINKING$>$
Step 1: Identify the skateboard as the primary object being ridden. The person is standing on it, indicating it is the platform for movement.
Step 2: Observe the wheel as a component of the skateboard. It is attached to the board’s underside and is part of the structural assembly that allows the skateboard to roll.
Step 3: Establish the relationship: the wheel is physically attached to the skateboard, and the skateboard is being used by the person. The wheel’s function is to support the skateboard’s movement on the ground, and it is inseparable from the skateboard’s structure.
Step 4: The interaction mechanism is mechanical: the wheel rolls on the pavement, enabling the skateboard to move. The skateboard contains the wheel as an integral part, and the person interacts with the skateboard via standing on it, which in turn causes the wheel to rotate and propel the skateboard forward.
Step 5: Therefore, the wheel is a part of the skateboard, and the skateboard is being used by the person to ride. The relationship is not of direct interaction between the wheel and the person, but rather a structural and functional relationship between the wheel and the skateboard, with the person using the skateboard.
$</$REASONING\_THINKING$>$

$<$FINAL\_SUMMARY$>$
wheel is part of skateboard
The wheel is an integral component of the skateboard, enabling its movement on the ground as the person rides it.
$</$FINAL\_SUMMARY$>$
    }{
        skateboard part of wheel
    }{
$<$VISUAL\_EVIDENCE$>$  
The image shows a skateboard positioned horizontally, with a wheel attached to its underside. The wheel is located near the front of the skateboard and appears to be connected to the deck via an axle, consistent with the typical mechanical structure of a skateboard. The wheel is aligned with the front edge of the deck, suggesting a functional integration rather than a decorative placement.
$</$VISUAL\_EVIDENCE$>$

$<$REASONING\_THINKING$>$  
The skateboard is a two-wheeled vehicle designed for motion on flat surfaces. The wheel is visibly connected to the underside of the deck, near the front, and its position and attachment style match the standard configuration of a skateboard’s front wheel assembly. This arrangement allows the wheel to rotate and engage with the ground, enabling forward movement. The spatial relationship indicates that the wheel is not merely placed nearby but is structurally integrated into the skateboard’s motion system, consistent with the function of a front wheel on a skateboard.
$</$REASONING\_THINKING$>$

$<$FINAL\_SUMMARY$>$  
The skateboard has a front wheel. The wheel is attached to the front of the skateboard and functions as part of its movement system.
$</$FINAL\_SUMMARY$>$
    }{
        skateboard has wheel 
    }

    \label{fig:qualitative_results}
\end{figure}

\begin{figure}[H]
    \centering
    
    % Sample 1
    \badcaseSample{
        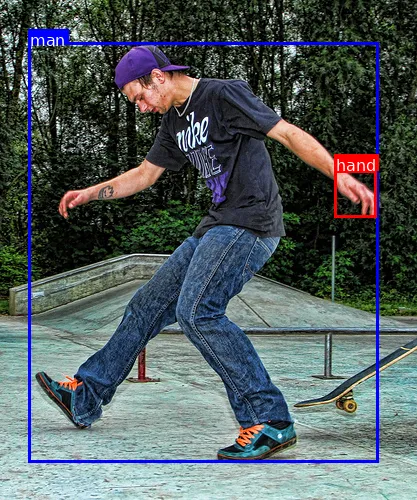 % 替换为你的图片路径
    }{$<$VISUAL\_EVIDENCE$>$  
The image shows a man positioned on a skateboard, with his body leaning forward and his arms slightly extended. The man's feet are placed on the deck of the skateboard, and the skateboard is oriented horizontally beneath him. The man appears to be actively engaged in a motion, possibly balancing or moving forward, with his upper body slightly bent. The overall posture suggests dynamic movement, and the man is clearly in control of the skateboard's direction and speed.
$</$VISUAL\_EVIDENCE$>$

$<$REASONING\_THINKING$>$  
The man is on the skateboard, which is a surface that allows for motion on the ground. The positioning of his feet and body indicates that he is using the skateboard as a means of transportation. His forward-leaning posture and extended arms suggest an effort to maintain balance while moving. The interaction between the man and the skateboard involves the man applying force to the deck with his feet, causing the skateboard to accelerate and navigate forward. This interaction is consistent with the man using the skateboard to propel himself forward and control the motion of the device.
$</$REASONING\_THINKING$>$

$<$FINAL\_SUMMARY$>$  
The relationship predicate is: man riding skateboard.  
The man is using the skateboard as a vehicle to move forward across the surface.
$</$FINAL\_SUMMARY$>$
    }{
        hand of man
    }{
        on
    }
    
    \label{fig:qualitative_results}
\end{figure}

\begin{figure}[H]
    \centering
    
    % Sample 1
    \badcaseSample{
    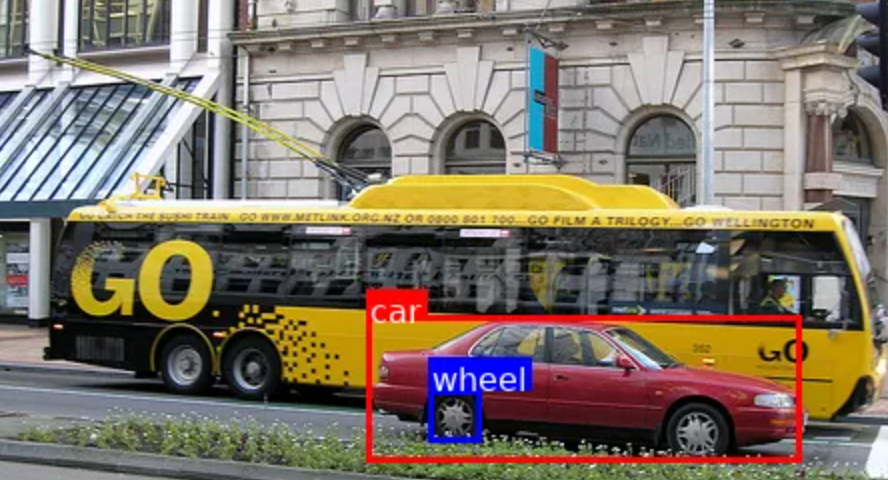 % 替换为你的图片路径
    }{$<$VISUAL\_EVIDENCE$>$  The image shows a car positioned horizontally, with a wheel located beneath it and slightly to the left. The wheel is smaller in size compared to the car and appears to be attached to the lower portion of the car's body. The wheel is aligned with the car's undercarriage, indicating a structural connection. The spatial arrangement suggests that the wheel is not independently placed but is instead integrated into the car's structure.$</$VISUAL\_EVIDENCE$>$ $<$REASONING\_THINKING$>$ 
The wheel is situated directly under the car and is oriented in a way that implies it is meant to rotate and support the car's movement. The wheel is connected to the car's chassis, which is typical of a wheel being mounted to a vehicle. The alignment and positioning suggest a functional integration where the wheel serves as a component that enables motion and support. This configuration indicates that the wheel is not simply adjacent to the car but is instead attached to it in a way that allows for movement and stability. The interaction mechanism involves the wheel being fixed to the car’s frame, enabling it to rotate and move the vehicle forward or backward.
$</$REASONING\_THINKING$>$
$<$FINAL\_SUMMARY$>$ 
car mounted on wheel  
The wheel is attached to the car's underside, enabling it to rotate and support the vehicle's movement.
$</$FINAL\_SUMMARY$>$
}{
        car mounted on wheel
    }{
        car has wheel
    }
    
    \label{fig:qualitative_results}
\end{figure}

\begin{figure}[H]
    \centering
    
    % Sample 1
    \badcaseSample{
        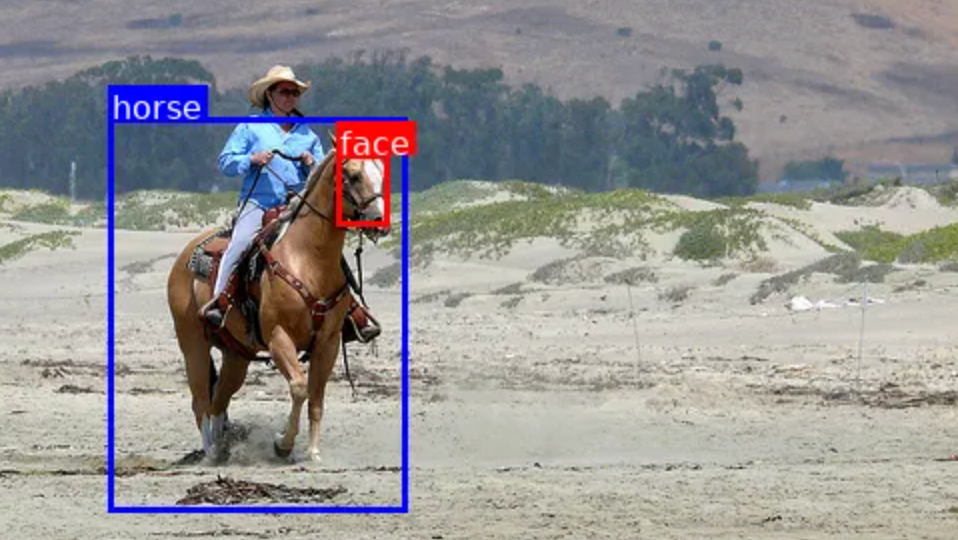 % 替换为你的图片路径
    }{$<$VISUAL\_EVIDENCE$>$ The image shows a horse with a rider positioned on its back. The rider is seated on the horse’s body, with their legs wrapped around the horse’s sides and their hands likely holding the reins. The horse is standing still, and the rider appears to be in a stable, upright posture, indicating control and balance. The positioning suggests a coordinated interaction between the rider and the horse.$</$VISUAL\_EVIDENCE$>$
    $<$REASONING\_THINKING$>$  
    The rider is located on the horse’s back, with their body aligned in a manner consistent with being seated. The proximity and posture of the rider suggest they are using the horse for transportation or control. The horse’s stable stance and the rider’s upright position imply a functional relationship where the rider is guiding or operating the horse. This coordination is typical of a controlled interaction involving movement or direction, which aligns with the idea of the rider being in control of the horse’s motion.$</$REASONING\_THINKING$>$
    $<$FINAL\_SUMMARY$>$ The relationship is: rider operates horse. The rider is seated on the horse and is actively controlling its movement.
    $</$FINAL\_SUMMARY$>$
    }{
        face riding horse
    }{
        on
    }
    \caption{Failure case of our method. The generated CoT contains hallucinated visual evidence and describes an incorrect object pair, leading to a mismatched relation prediction.}
    
    \label{fig:qualitative_results}
\end{figure}

%% file: sections/checklist.tex
\clearpage
\section*{NeurIPS Paper Checklist}

\begin{enumerate}

\item {\bf Claims}
    \item[] Question: Do the main claims made in the abstract and introduction accurately reflect the paper's contributions and scope?
    \item[] Answer: \answerYes{} % Replace by \answerYes{}, \answerNo{}, or \answerNA{}.
    \item[] Justification: We reported the claim in the abstract and introduction.

\item {\bf Limitations}
    \item[] Question: Does the paper discuss the limitations of the work performed by the authors?
    \item[] Answer: \answerYes{} % Replace by \answerYes{}, \answerNo{}, or \answerNA{}.
    \item[] Justification: We reported the limitation as the end of paper.

\item {\bf Theory assumptions and proofs}
    \item[] Question: For each theoretical result, does the paper provide the full set of assumptions and a complete (and correct) proof?
    \item[] Answer: \answerNA{} % Replace by \answerYes{}, \answerNo{}, or \answerNA{}.
    \item[] Justification: The paper is not about theory.

    \item {\bf Experimental result reproducibility}
    \item[] Question: Does the paper fully disclose all the information needed to reproduce the main experimental results of the paper to the extent that it affects the main claims and/or conclusions of the paper (regardless of whether the code and data are provided or not)?
    \item[] Answer: \answerYes % Replace by \answerYes{}, \answerNo{}, or \answerNA{}.
    \item[] Justification: We provided details about the methodology and implementation in the main paper and appendix.

\item {\bf Open access to data and code}
    \item[] Question: Does the paper provide open access to the data and code, with sufficient instructions to faithfully reproduce the main experimental results, as described in supplemental material?
    \item[] Answer: \answerYes{} % Replace by \answerYes{}, \answerNo{}, or \answerNA{}.
    \item[] Justification: The code and model will be released.

\item {\bf Experimental setting/details}
    \item[] Question: Does the paper specify all the training and test details (e.g., data splits, hyperparameters, how they were chosen, type of optimizer) necessary to understand the results?
    \item[] Answer: \answerYes{} % Replace by \answerYes{}, \answerNo{}, or \answerNA{}.
    \item[] Justification: We reported the experimental setup and details in the main paper and appendix

\item {\bf Experiment statistical significance}
    \item[] Question: Does the paper report error bars suitably and correctly defined or other appropriate information about the statistical significance of the experiments?
    \item[] Answer: \answerNo{} % Replace by \answerYes{}, \answerNo{}, or \answerNA{}.
    \item[] Justification: Due to limited computational resources, we were unable to run the experiments multiple times to compute error bars.

\item {\bf Experiments compute resources}
    \item[] Question: For each experiment, does the paper provide sufficient information on the computer resources (type of compute workers, memory, time of execution) needed to reproduce the experiments?
    \item[] Answer: \answerYes{} % Replace by \answerYes{}, \answerNo{}, or \answerNA{}.
    \item[] Justification: Its reported in the paper and appendix.
    
\item {\bf Code of ethics}
    \item[] Question: Does the research conducted in the paper conform, in every respect, with the NeurIPS Code of Ethics \url{https://neurips.cc/public/EthicsGuidelines}?
    \item[] Answer: \answerYes{} % Replace by \answerYes{}, \answerNo{}, or \answerNA{}.
    \item[] Justification: We have carefully reviewed the NeurIPS Code of Ethics and confirm that the research conducted and presented in this paper fully complies with all ethical guidelines.

\item {\bf Broader impacts}
    \item[] Question: Does the paper discuss both potential positive societal impacts and negative societal impacts of the work performed?
    \item[] Answer: \answerYes{} % Replace by \answerYes{}, \answerNo{}, or \answerNA{}.
    \item[] Justification: Section~5 briefly discusses potential benefits and downstream risks.
    
\item {\bf Safeguards}
    \item[] Question: Does the paper describe safeguards that have been put in place for responsible release of data or models that have a high risk for misuse (e.g., pre-trained language models, image generators, or scraped datasets)?
    \item[] Answer: \answerNA{} % Replace by \answerYes{}, \answerNo{}, or \answerNA{}.
    \item[] Justification: The paper poses no such risks.

\item {\bf Licenses for existing assets}
    \item[] Question: Are the creators or original owners of assets (e.g., code, data, models), used in the paper, properly credited and are the license and terms of use explicitly mentioned and properly respected?
    \item[] Answer: \answerYes{} % Replace by \answerYes{}, \answerNo{}, or \answerNA{}.
    \item[] Justification: We cite the datasets, pretrained models, and third-party code used in this work and have checked their licenses and complied with the applicable terms of use.

\item {\bf New assets}
    \item[] Question: Are new assets introduced in the paper well documented and is the documentation provided alongside the assets?
    \item[] Answer: \answerNA{} % Replace by \answerYes{}, \answerNo{}, or \answerNA{}.
    \item[] Justification: The paper did not introduce a new asset.

\item {\bf Crowdsourcing and research with human subjects}
    \item[] Question: For crowdsourcing experiments and research with human subjects, does the paper include the full text of instructions given to participants and screenshots, if applicable, as well as details about compensation (if any)? 
    \item[] Answer: \answerNA{} % Replace by \answerYes{}, \answerNo{}, or \answerNA{}.
    \item[] Justification: The paper does not involve crowdsourcing nor research with human subjects.

\item {\bf Institutional review board (IRB) approvals or equivalent for research with human subjects}
    \item[] Question: Does the paper describe potential risks incurred by study participants, whether such risks were disclosed to the subjects, and whether Institutional Review Board (IRB) approvals (or an equivalent approval/review based on the requirements of your country or institution) were obtained?
    \item[] Answer: \answerNA{} % Replace by \answerYes{}, \answerNo{}, or \answerNA{}.
    \item[] Justification: The paper does not involve crowdsourcing nor research with human subjects.


\item {\bf Declaration of LLM usage}
    \item[] Question: Does the paper describe the usage of LLMs if it is an important, original, or non-standard component of the core methods in this research? Note that if the LLM is used only for writing, editing, or formatting purposes and does \emph{not} impact the core methodology, scientific rigor, or originality of the research, declaration is not required.
    %this research? 
    \item[] Answer: \answerYes{} % Replace by \answerYes{}, \answerNo{}, or \answerNA{}.
    \item[] Justification: The MLLM is the core component of our framework. We have detailed the specific MLLMs used, the prompting strategies, and the generation pipeline in the paper.

\end{enumerate}

%% file: neurips_2026.bbl
\begin{thebibliography}{10}

\bibitem{bai2025qwen3}
Shuai Bai, Yuxuan Cai, Ruizhe Chen, Keqin Chen, Xionghui Chen, Zesen Cheng, Lianghao Deng, Wei Ding, Chang Gao, Chunjiang Ge, et~al.
\newblock Qwen3-vl technical report.
\newblock {\em arXiv preprint arXiv:2511.21631}, 2025.

\bibitem{chao2018learning}
Yu-Wei Chao, Yunfan Liu, Xieyang Liu, Huayi Zeng, and Jia Deng.
\newblock Learning to detect human-object interactions.
\newblock In {\em 2018 ieee winter conference on applications of computer vision (wacv)}, pages 381--389. IEEE, 2018.

\bibitem{chen2020simpleframeworkcontrastivelearning}
Ting Chen, Simon Kornblith, Mohammad Norouzi, and Geoffrey Hinton.
\newblock A simple framework for contrastive learning of visual representations, 2020.

\bibitem{chen2024expanding}
Zuyao Chen, Jinlin Wu, Zhen Lei, Zhaoxiang Zhang, and Chang~Wen Chen.
\newblock Expanding scene graph boundaries: Fully open-vocabulary scene graph generation via visual-concept alignment and retention.
\newblock In {\em European Conference on Computer Vision}, pages 108--124. Springer, 2024.

\bibitem{dutta2025open}
Amartya Dutta, Kazi~Sajeed Mehrab, Medha Sawhney, Abhilash Neog, Mridul Khurana, Sepideh Fatemi, Aanish Pradhan, M~Maruf, Ismini Lourentzou, Arka Daw, et~al.
\newblock Open world scene graph generation using vision language models.
\newblock {\em arXiv preprint arXiv:2506.08189}, 2025.

\bibitem{gao2021simcse}
Tianyu Gao, Xingcheng Yao, and Danqi Chen.
\newblock Simcse: Simple contrastive learning of sentence embeddings.
\newblock In {\em Proceedings of the 2021 conference on empirical methods in natural language processing}, pages 6894--6910, 2021.

\bibitem{gu2019unpaired}
Jiuxiang Gu, Shafiq Joty, Jianfei Cai, Handong Zhao, Xu~Yang, and Gang Wang.
\newblock Unpaired image captioning via scene graph alignments.
\newblock In {\em Proceedings of the IEEE/CVF international conference on computer vision}, pages 10323--10332, 2019.

\bibitem{Guo_2025}
Daya Guo, Dejian Yang, Haowei Zhang, et~al.
\newblock Deepseek-r1 incentivizes reasoning in llms through reinforcement learning.
\newblock {\em Nature}, 645(8081):633–638, 2025.

\bibitem{haochen2021provable}
Jeff~Z HaoChen, Colin Wei, Adrien Gaidon, and Tengyu Ma.
\newblock Provable guarantees for self-supervised deep learning with spectral contrastive loss.
\newblock {\em Advances in neural information processing systems}, 34:5000--5011, 2021.

\bibitem{he2020momentumcontrastunsupervisedvisual}
Kaiming He, Haoqi Fan, Yuxin Wu, Saining Xie, and Ross Girshick.
\newblock Momentum contrast for unsupervised visual representation learning, 2020.

\bibitem{he2022towards}
Tao He, Lianli Gao, Jingkuan Song, and Yuan-Fang Li.
\newblock Towards open-vocabulary scene graph generation with prompt-based finetuning.
\newblock In {\em European conference on computer vision}, pages 56--73. Springer, 2022.

\bibitem{hudson2019gqa}
Drew~A Hudson and Christopher~D Manning.
\newblock Gqa: A new dataset for real-world visual reasoning and compositional question answering.
\newblock In {\em Proceedings of the IEEE/CVF conference on computer vision and pattern recognition}, pages 6700--6709, 2019.

\bibitem{Johnson_2015_CVPR}
Justin Johnson, Ranjay Krishna, Michael Stark, Li-Jia Li, David Shamma, Michael Bernstein, and Li~Fei-Fei.
\newblock Image retrieval using scene graphs.
\newblock In {\em Proceedings of the IEEE Conference on Computer Vision and Pattern Recognition (CVPR)}, June 2015.

\bibitem{kim2024llm4sgglargelanguagemodels}
Kibum Kim, Kanghoon Yoon, Jaehyeong Jeon, Yeonjun In, Jinyoung Moon, Donghyun Kim, and Chanyoung Park.
\newblock Llm4sgg: Large language models for weakly supervised scene graph generation, 2024.

\bibitem{openSGen}
Zihan Kong and Haiwei Zhang.
\newblock Opensgen: Fine-grained relation-aware prompt for open-vocabulary scene graph generation.
\newblock In {\em Proceedings of the 2025 International Conference on Multimedia Retrieval}, ICMR '25, page 634–643, New York, NY, USA, 2025. Association for Computing Machinery.

\bibitem{krishna2017visual}
Ranjay Krishna, Yuke Zhu, Oliver Groth, Justin Johnson, Kenji Hata, Joshua Kravitz, Stephanie Chen, Yannis Kalantidis, Li-Jia Li, David~A Shamma, et~al.
\newblock Visual genome: Connecting language and vision using crowdsourced dense image annotations.
\newblock {\em International journal of computer vision}, 123(1):32--73, 2017.

\bibitem{lei2023efficient}
Ting Lei, Fabian Caba, Qingchao Chen, Hailin Jin, Yuxin Peng, and Yang Liu.
\newblock Efficient adaptive human-object interaction detection with concept-guided memory.
\newblock In {\em Proceedings of the IEEE/CVF international conference on computer vision}, pages 6480--6490, 2023.

\bibitem{li2025relationr1progressivelycognitivechainofthought}
Lin Li, Wei Chen, Jiahui Li, Kwang-Ting Cheng, and Long Chen.
\newblock Relation-r1: Progressively cognitive chain-of-thought guided reinforcement learning for unified relation comprehension, 2025.

\bibitem{li2025interaction}
Lin Li, Chuhan Zhang, Dong Zhang, Chong Sun, Chen Li, and Long Chen.
\newblock Interaction-centric knowledge infusion and transfer for open-vocabulary scene graph generation.
\newblock {\em arXiv preprint arXiv:2511.05935}, 2025.

\bibitem{li2026qwen3}
Mingxin Li, Yanzhao Zhang, Dingkun Long, Keqin Chen, Sibo Song, Shuai Bai, Zhibo Yang, Pengjun Xie, An~Yang, Dayiheng Liu, et~al.
\newblock Qwen3-vl-embedding and qwen3-vl-reranker: A unified framework for state-of-the-art multimodal retrieval and ranking.
\newblock {\em arXiv preprint arXiv:2601.04720}, 2026.

\bibitem{li2024sgtrendtoendscenegraph}
Rongjie Li, Songyang Zhang, and Xuming He.
\newblock Sgtr+: End-to-end scene graph generation with transformer, 2024.

\bibitem{li2024pixelsgraphsopenvocabularyscene}
Rongjie Li, Songyang Zhang, Dahua Lin, Kai Chen, and Xuming He.
\newblock From pixels to graphs: Open-vocabulary scene graph generation with vision-language models, 2024.

\bibitem{liu2025relationawarehierarchicalpromptopenvocabulary}
Tao Liu, Rongjie Li, Chongyu Wang, and Xuming He.
\newblock Relation-aware hierarchical prompt for open-vocabulary scene graph generation, 2025.

\bibitem{ouyang2022traininglanguagemodelsfollow}
Long Ouyang, Jeff Wu, Xu~Jiang, Diogo Almeida, Carroll~L. Wainwright, Pamela Mishkin, Chong Zhang, Sandhini Agarwal, Katarina Slama, Alex Ray, John Schulman, Jacob Hilton, Fraser Kelton, Luke Miller, Maddie Simens, Amanda Askell, Peter Welinder, Paul Christiano, Jan Leike, and Ryan Lowe.
\newblock Training language models to follow instructions with human feedback, 2022.

\bibitem{radford2021learningtransferablevisualmodels}
Alec Radford, Jong~Wook Kim, Chris Hallacy, Aditya Ramesh, Gabriel Goh, Sandhini Agarwal, Girish Sastry, Amanda Askell, Pamela Mishkin, Jack Clark, Gretchen Krueger, and Ilya Sutskever.
\newblock Learning transferable visual models from natural language supervision, 2021.

\bibitem{shao2024deepseekmathpushinglimitsmathematical}
Zhihong Shao, Peiyi Wang, Qihao Zhu, Runxin Xu, Junxiao Song, Xiao Bi, Haowei Zhang, Mingchuan Zhang, Y.~K. Li, Y.~Wu, and Daya Guo.
\newblock Deepseekmath: Pushing the limits of mathematical reasoning in open language models, 2024.

\bibitem{shi2019explainable}
Jiaxin Shi, Hanwang Zhang, and Juanzi Li.
\newblock Explainable and explicit visual reasoning over scene graphs.
\newblock In {\em Proceedings of the IEEE/CVF conference on computer vision and pattern recognition}, pages 8376--8384, 2019.

\bibitem{Tang_2020_CVPR}
Kaihua Tang, Yulei Niu, Jianqiang Huang, Jiaxin Shi, and Hanwang Zhang.
\newblock Unbiased scene graph generation from biased training.
\newblock In {\em Proceedings of the IEEE/CVF Conference on Computer Vision and Pattern Recognition (CVPR)}, June 2020.

\bibitem{oord2019representationlearningcontrastivepredictive}
Aaron van~den Oord, Yazhe Li, and Oriol Vinyals.
\newblock Representation learning with contrastive predictive coding, 2019.

\bibitem{wang2020understanding}
Tongzhou Wang and Phillip Isola.
\newblock Understanding contrastive representation learning through alignment and uniformity on the hypersphere.
\newblock In {\em International conference on machine learning}, pages 9929--9939. PMLR, 2020.

\bibitem{wang2018scene}
Yu-Siang Wang, Chenxi Liu, Xiaohui Zeng, and Alan Yuille.
\newblock Scene graph parsing as dependency parsing.
\newblock In {\em Proceedings of the 2018 Conference of the North American Chapter of the Association for Computational Linguistics: Human Language Technologies, Volume 1 (Long Papers)}, pages 397--407, 2018.

\bibitem{wei2023chainofthoughtpromptingelicitsreasoning}
Jason Wei, Xuezhi Wang, Dale Schuurmans, Maarten Bosma, Brian Ichter, Fei Xia, Ed~Chi, Quoc Le, and Denny Zhou.
\newblock Chain-of-thought prompting elicits reasoning in large language models, 2023.

\bibitem{xu2017scene}
Danfei Xu, Yuke Zhu, Christopher~B Choy, and Li~Fei-Fei.
\newblock Scene graph generation by iterative message passing.
\newblock In {\em Proceedings of the IEEE conference on computer vision and pattern recognition}, pages 5410--5419, 2017.

\bibitem{xu2024llavaspacesggvisualinstructtuning}
Mingjie Xu, Mengyang Wu, Yuzhi Zhao, Jason Chun~Lok Li, and Weifeng Ou.
\newblock Llava-spacesgg: Visual instruct tuning for open-vocabulary scene graph generation with enhanced spatial relations, 2024.

\bibitem{xuan2026zero}
Shiyu Xuan, Dongkai Wang, Zechao Li, and Jinhui Tang.
\newblock Zero-shot hoi detection with mllm-based detector-agnostic interaction recognition.
\newblock {\em arXiv preprint arXiv:2602.15124}, 2026.

\bibitem{yang2022panoptic}
Jingkang Yang, Yi~Zhe Ang, Zujin Guo, Kaiyang Zhou, Wayne Zhang, and Ziwei Liu.
\newblock Panoptic scene graph generation.
\newblock In {\em European conference on computer vision}, pages 178--196. Springer, 2022.

\bibitem{yang2019auto}
Xu~Yang, Kaihua Tang, Hanwang Zhang, and Jianfei Cai.
\newblock Auto-encoding scene graphs for image captioning.
\newblock In {\em Proceedings of the IEEE/CVF conference on computer vision and pattern recognition}, pages 10685--10694, 2019.

\bibitem{yu2023visuallyEPIC}
Qifan Yu, Juncheng Li, Yu~Wu, Siliang Tang, Wei Ji, and Yueting Zhuang.
\newblock Visually-prompted language model for fine-grained scene graph generation in an open world.
\newblock In {\em Proceedings of the IEEE/CVF International Conference on Computer Vision}, pages 21560--21571, 2023.

\bibitem{Zellers_2018_CVPR}
Rowan Zellers, Mark Yatskar, Sam Thomson, and Yejin Choi.
\newblock Neural motifs: Scene graph parsing with global context.
\newblock In {\em Proceedings of the IEEE Conference on Computer Vision and Pattern Recognition (CVPR)}, June 2018.

\bibitem{zhang2022efficient}
Frederic~Z Zhang, Dylan Campbell, and Stephen Gould.
\newblock Efficient two-stage detection of human-object interactions with a novel unary-pairwise transformer.
\newblock In {\em 2022 IEEE/CVF Conference on Computer Vision and Pattern Recognition (CVPR)}, pages 20072--20080. IEEE, 2022.

\bibitem{VS3}
Yong Zhang, Yingwei Pan, Ting Yao, Rui Huang, Tao Mei, and Chang-Wen Chen.
\newblock Learning to generate language-supervised and open-vocabulary scene graph using pre-trained visual-semantic space.
\newblock In {\em 2023 IEEE/CVF Conference on Computer Vision and Pattern Recognition (CVPR)}, pages 2915--2924, 2023.

\bibitem{zhang2024multimodalchainofthoughtreasoninglanguage}
Zhuosheng Zhang, Aston Zhang, Mu~Li, Hai Zhao, George Karypis, and Alex Smola.
\newblock Multimodal chain-of-thought reasoning in language models, 2024.

\bibitem{zheng2023prototypebasedembeddingnetworkscene}
Chaofan Zheng, Xinyu Lyu, Lianli Gao, Bo~Dai, and Jingkuan Song.
\newblock Prototype-based embedding network for scene graph generation, 2023.

\bibitem{zhou2024openpsgopensetpanopticscene}
Zijian Zhou, Zheng Zhu, Holger Caesar, and Miaojing Shi.
\newblock Openpsg: Open-set panoptic scene graph generation via large multimodal models, 2024.

\end{thebibliography}
